\documentclass{article}

\usepackage{guanxing}

\usepackage{amsmath,amsfonts,bm}

\def\eqref#1{equation~\ref{#1}}

\def\1{\bm{1}}

\DeclareMathAlphabet{\mathsfit}{\encodingdefault}{\sfdefault}{m}{sl}
\SetMathAlphabet{\mathsfit}{bold}{\encodingdefault}{\sfdefault}{bx}{n}

\usepackage{hyperref}
\usepackage{url}
\usepackage{capt-of}
\usepackage{wrapfig}
\usepackage{needspace}
\usepackage{booktabs}
\usepackage{amssymb}
\usepackage{xcolor,graphicx,adjustbox,placeins,etoolbox}

\title{Proactive Physical AI through Continuous Observation and Timely Intervention}

\author{\mbox{Zhenhua Yang\textsuperscript{1}}, \mbox{Jiafeng Liang\textsuperscript{1}}, \mbox{Xingyu Zhou\textsuperscript{1}}, \mbox{Yi Jiang\textsuperscript{1}}, \mbox{Xiayu Cao\textsuperscript{1}}, \mbox{Zui Chen\textsuperscript{1}}, \mbox{Luqi Wang\textsuperscript{1}}, \mbox{Haiyang Sun\textsuperscript{1}}, \mbox{Yufei Xiang\textsuperscript{1}}, \mbox{Chang Li\textsuperscript{1}}, \mbox{Rongchuan Mu\textsuperscript{1}}, \mbox{Bowen Chi\textsuperscript{1}}, \mbox{Zihe Tong\textsuperscript{1}}, \mbox{Mingyu Li\textsuperscript{1}}, \mbox{Huijie Ma\textsuperscript{1}}, \mbox{Yiqiao Yang\textsuperscript{1}}, \mbox{Baoqi Ren\textsuperscript{1}}, \mbox{Zhiming Ma\textsuperscript{1}}, \mbox{Zhihao Zhu\textsuperscript{1}}, \mbox{Jiahao Ji\textsuperscript{1}}, \mbox{Zhuoyao Wang\textsuperscript{1}}, \mbox{Qianyu Wang\textsuperscript{1}}, \mbox{Xiangyu Liu\textsuperscript{1}}, \mbox{Wang Zixiang\textsuperscript{1}}, \mbox{Shunliang Zhang\textsuperscript{1}}, \mbox{Ming Liu\textsuperscript{1}}, \mbox{Xiaocheng Feng\textsuperscript{1}}, \mbox{Duyu Tang\textsuperscript{2}}, \mbox{Bing Qin\textsuperscript{1}}}
\gxshorttitle{Proactive Physical AI}
\gxaffiliation{\textsuperscript{1}Research Center for Social Computing and Interactive Robotics, Harbin Institute of Technology\\[3pt]\textsuperscript{2}Guanxing Technology}
\gxseries{RESEARCH PAPER}

\usepackage{xcolor}
\usepackage{booktabs}
\usepackage{array}
\usepackage{amsmath,amssymb}
\usepackage{tcolorbox}
\tcbuselibrary{skins,breakable}

\definecolor{ProPAIChallengeBlue}{HTML}{B9471D}
\definecolor{ProPAIChallengePale}{HTML}{FFF9F5}
\definecolor{ProPAIChallengeBorder}{HTML}{EBC4B5}
\definecolor{ProPAIChallengeText}{HTML}{343D4B}

\newtcolorbox[auto counter]{propaopenchallenge}[1][]{
  enhanced,
  breakable,
  colback=ProPAIChallengePale,
  colframe=ProPAIChallengeBorder,
  colbacktitle=ProPAIChallengeBlue,
  coltitle=white,
  coltext=ProPAIChallengeText,
  fonttitle=\bfseries,
  fontupper=\itshape,
  title={Open Challenge~\thetcbcounter},
  boxrule=0.5pt,
  arc=1mm,
  left=3mm,
  right=3mm,
  top=2mm,
  bottom=2mm,
  toptitle=1.5mm,
  bottomtitle=1.5mm,
  before skip=10pt,
  after skip=8pt,
  #1
}

\begin{document}

\maketitle
\begingroup
\setlength{\textfloatsep}{10pt plus 2pt minus 2pt}
\setlength{\floatsep}{8pt plus 2pt minus 2pt}
\setlength{\intextsep}{10pt plus 2pt minus 2pt}

\begin{abstract}

Physical AI is commonly formulated around an explicitly specified task: given a goal or instruction, a system perceives, reasons, and acts. In many real-world settings, however, important tasks emerge from continuously changing world states before a person explicitly formulates them. We study this complementary problem as {Proactive Physical AI}, where an AI system continuously observes the physical world, identifies situations that warrant intervention, and initiates appropriate actions at the right time.
We develop {ProAgent}, an end-to-end system built around two components: a Jev-style proactive timing model (dubbed StreamingJev) that determines whether and when intervention is needed, and a Response Model that generates responses for people or invokes available device capabilities for physical execution. To support learning at scale, we develop an automated data pipeline that converts videos into supervision for proactive timing. Experiments demonstrate strong proactive performance under both explicit objectives and standing roles, combining more accurate intervention timing with fewer false triggers than the source-backbone baseline. Real-robot evaluations further demonstrate ProAgent's ability to identify emerging needs and translate them into appropriate physical actions.
\end{abstract}

\begingroup
\colorlet{blue}{black}
\begin{minipage}{\linewidth}\captionsetup{hypcap=false}
    \centering
    \includegraphics[width=.95\linewidth]{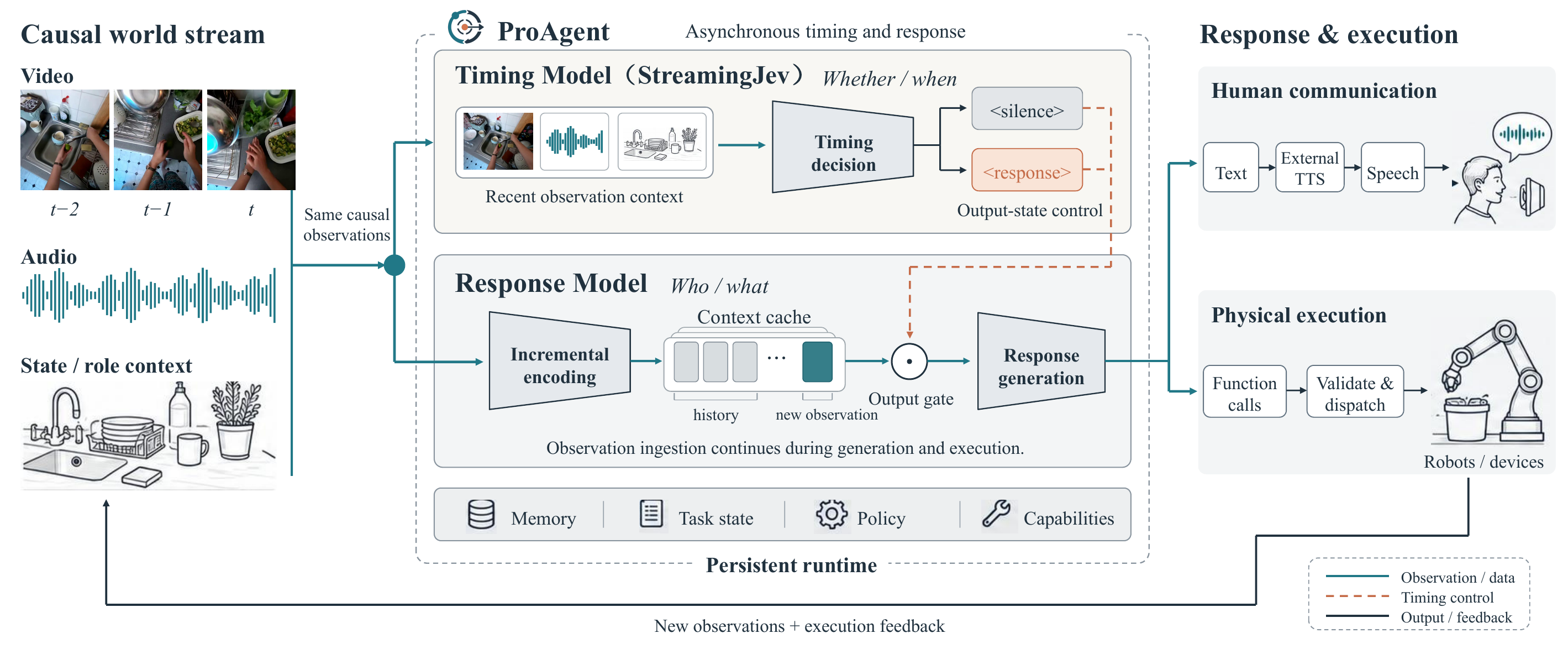}
    {\color{blue}
    \captionof{figure}{\textbf{ProAgent architecture.} Both models consume the causal world stream asynchronously; the Response Model can accumulate context while the Timing Model (StreamingJev) remains silent. A response decision enables output generation, not the ingestion of observations. Human-facing text is passed to external TTS. Physical outputs use typed function calls, while runtime validation and dispatch connect them to device providers. Feedback returns to the continuing observation loop.}
    \label{fig:propa-engine}}
\end{minipage}

\endgroup
\clearpage

\section{Introduction}
\label{sec:introduction}

Artificial intelligence systems are increasingly capable of acting in the physical world, yet they are still commonly formulated around an explicitly specified task~\citep{ahn2022saycan,brohan2023rt2,physicalintelligence2025pi05}. A robot may be asked to pick up an object, a household assistant to operate an appliance, or a multimodal model to answer a question about its surroundings. In practice, however, many useful tasks arise before anyone explicitly asks for them. Consider a kitchen assistant observing a pot on a stove: as the liquid begins to boil over, the important problem is not how to execute an already specified instruction, but whether the system can recognize that intervention is becoming necessary, determine when to act, and take an appropriate response before the opportunity is lost. Similar situations arise when a person is about to leave an important object behind, when a worker approaches a hazardous area, or when an appliance enters an abnormal state. In such cases, intelligence begins not from a human-issued task, but from continuously changing world states. Proactivity therefore changes not only how quickly a physical AI system responds, but also \emph{where tasks originate}. Prior work on proactive human--robot interaction and context-aware assistants motivates studying initiative beyond explicit task requests~\citep{vandenbroek2024proactivehri,wang2025roboomni,yang2025proagent}.

We study this setting as \textbf{Proactive Physical AI}. Rather than assuming that a task has already been recognized and specified, a ProPAI system continuously observes the physical world and determines whether the evolving situation calls for intervention. When intervention is warranted, the system must infer what needs to be done, decide when to act, and select an appropriate response for the available executor, which may be a person, robot, or other physical device. This formulation introduces three tightly coupled challenges. First, the system must continuously interpret multimodal observations and identify changes that are relevant to its responsibilities. Second, it must reason about {intervention timing}: acting too early may be unnecessary or disruptive, whereas acting too late may miss the useful window altogether. Third, the inferred need must be grounded in the capabilities and constraints of available executors so that an intended response can be translated into an actionable physical operation. Together, these challenges extend physical AI from executing a given task to identifying and addressing tasks as they emerge from the world.

To investigate this problem, we develop \textbf{ProAgent}, an end-to-end system for continuous observation, proactive decision making, and physical execution. ProAgent separates proactive decision making into two complementary components. A Jev-style proactive timing model, dubbed \textbf{StreamingJev}, continuously processes streaming observations and determines whether the current situation warrants intervention, while a {Response Model} reasons over the observed context to determine whom to address and what action or communication should follow. To train proactive timing behavior at scale, we introduce an automated data pipeline that converts Internet videos into supervision for both intervention and non-intervention moments, producing 224,268 causal decision sequences from 20,258 source videos without requiring task instructions to be manually specified for every scene. The two models operate asynchronously, allowing environmental observation to continue while responses are being generated or executed. When the response targets a person, the system generates a communicative response; when it targets a device or robot, the system invokes the capabilities exposed by the corresponding executor. This design provides a practical mechanism for connecting continuous world observation to timely and capability-grounded action.

Experiments demonstrate that ProAgent outperforms the evaluated standalone multimodal models and streaming assistants in intervention timing and time-aware response quality. Under explicit assistance objectives, ProAgent achieves an event-level F1 of 42.43\%, exceeding the strongest evaluated baseline by 6.27 percentage points, while attaining the highest PAUC under both explicit objectives and standing roles. We further evaluate ProAgent in real robotic scenarios with Reachy Mini, demonstrating its ability to identify emerging needs, intervene at appropriate moments, and execute corresponding physical actions.
\section{Problem Formulation}
\label{sec:formulation}

We consider a physical AI system operating in a continuously evolving environment. Conventional task-conditioned physical AI assumes that an explicit task is provided before execution, and the system uses observations of the world to determine how to accomplish that task. \textbf{Proactive Physical AI (ProPAI)} removes this assumption: no concrete task is specified in advance. Instead, given continuous observations of the physical world and a standing context that defines its responsibilities and constraints, the system must determine \emph{whether} the current situation warrants intervention, \emph{when} to intervene, and \emph{what} response can be appropriately executed. Task initiation is therefore part of the inference problem rather than a prerequisite for it.

\paragraph{Input.}
At each time step $t$, the system receives
\begin{equation}
    \mathcal I_t =
    (\mathbf{o}_t,\mathcal H_t,\mathcal D_t,\mathbf f_t),
\end{equation}
where $\mathbf{o}_t$ denotes the current multimodal observations, such as vision and audio; $\mathcal H_t$ denotes the people currently involved; $\mathcal D_t$ denotes the available physical devices; and $\mathbf f_t$ denotes feedback from previous interactions or executions. Each device $d\in\mathcal D_t$ exposes a set of currently available capabilities $\mathcal A_t(d)$.

In addition, the system is provided with a standing context $\mathcal C$ specifying its role, permissions, and operating constraints. Importantly, $\mathcal C$ defines the \emph{scope} within which the system may assist, but does not specify which event will require intervention, what concrete task should be performed, or when that task should begin. At time $t$, the system makes decisions only from the observed history
\begin{equation}
    \mathcal I_{\leq t}
    =(\mathcal I_1,\ldots,\mathcal I_t)
\end{equation}
and $\mathcal C$, without access to future observations.

\paragraph{Output.}
At every time step, the system must first decide whether intervention is warranted. We formulate the policy as
\begin{equation}
    (z_t,\mathcal U_t)
    =
    \pi(\mathcal I_{\leq t},\mathcal C),
\end{equation}
where $z_t\in\{0,1\}$ is the intervention decision. When $z_t=0$, the system remains silent and continues observing. When $z_t=1$, $\mathcal U_t$ contains one or more responses selected from the currently feasible response space
\begin{equation}
\begin{aligned}
\mathcal R_t
=&\;
\{
\texttt{COMMUNICATE}(h,m)
\,|\,
h\in\mathcal H_t
\}\\
&\cup
\{
\texttt{ACT}(d,a)
\,|\,
d\in\mathcal D_t,\,
a\in\mathcal A_t(d)
\}.
\end{aligned}
\end{equation}
Here, $\texttt{COMMUNICATE}(h,m)$ delivers a message $m$ to a person $h$, while $\texttt{ACT}(d,a)$ invokes capability $a$ on device $d$. An intervention may contain multiple responses, allowing communication and physical actions to be coordinated when necessary.
Equivalently, the observable system output is
\begin{equation}
    \mathcal Y_t =
    \begin{cases}
        \texttt{SILENT}, & z_t=0,\\
        \mathcal U_t\subseteq\mathcal R_t, & z_t=1.
    \end{cases}
\end{equation}
Because this decision is repeatedly evaluated over the observation stream, \emph{when} the policy first changes from continued observation to intervention is itself part of the prediction problem. Remaining silent is therefore a first-class decision rather than the absence of a prediction.

\paragraph{From Passive to Proactive Physical AI.}
% For precision, we use \emph{task-conditioned physical AI} to describe the setting we informally refer to as passive physical AI. 
The distinction between passive and proactive systems is not whether a system can perceive its environment, autonomously plan, or use execution feedback. The distinction is instead \textbf{where the task comes from}.

In the passive setting, a concrete task $q$ is supplied externally before execution:
\begin{equation}
    \mathcal Y_t^{\mathrm{passive}}
    =
    \pi_{\mathrm{passive}}
    (\mathcal I_{\leq t},\mathcal C,q).
\end{equation}
World observations are used to determine how to accomplish $q$. In ProPAI, however, no such task variable is provided:
\begin{equation}
    \mathcal Y_t^{\mathrm{proactive}}
    =
    \pi_{\mathrm{proactive}}
    (\mathcal I_{\leq t},\mathcal C).
\end{equation}
The system must instead infer from the evolving world state whether any task should be initiated, determine an appropriate intervention time, and select a feasible response. Conceptually, the two paradigms can be summarized as
\begin{equation}
\begin{aligned}
\text{Task-conditioned Physical AI:}\quad
&\text{Task} + \text{World}
\rightarrow
\text{Action},\\
\text{Proactive Physical AI:}\quad
&\text{World Stream}
\rightarrow
\text{Intervention Decision}
\rightarrow
\text{Action}.
\end{aligned}
\end{equation}

ProPAI introduces an additional decision layer before execution: determining whether a task exists, when it should be initiated, and what intervention is appropriate from continuous observations.

\section{A Proactive Physical AI System}
\label{sec:system}

Proactive assistance requires frequent decisions about whether to intervene,
while generating a response or carrying out an action can take much longer.
ProAgent therefore separates timing prediction from response generation,
allowing observation to continue while an intervention is underway.
As shown in Figure~\ref{fig:propa-engine}, the Timing Model (StreamingJev) determines
\emph{when to intervene}, and the Response Model determines \emph{whom to
address and what to do}. A persistent runtime connects these decisions
to the evolving world state and the feedback from earlier actions.

Both models receive causal observations together with standing roles,
available device capabilities, and interaction history. While the Timing
Model evaluates whether a response is warranted, the Response Model
incrementally encodes the incoming context. A trigger enables generation
from this prepared context, so the system need not wait until intervention
is required to begin processing the observations. Observation ingestion
continues during generation and execution. The runtime tracks responses
and operations already in progress, allowing subsequent timing decisions
to account for an ongoing intervention instead of starting a new request
at every update. The models maintain separate caches for their respective
weights and inputs; shared observations do not require shared KV states.

\label{sec:response-model}
\label{sec:pci}
\label{sec:action}
The Response Model is a multimodal foundation model conditioned
on this context and the available capabilities. It produces text for external TTS when addressing a person, or
function calls when directing a device. The runtime validates and
dispatches these calls, then returns execution results to the interaction
state. Together with subsequent observations, this feedback allows the
system to determine whether the need has been addressed or further
assistance is required. The result is a continuing observation--intervention
loop rather than a sequence of independent responses. 
Detailed state
management, cache validity, and executor-supported cancellation and
recovery are described in Appendix~\ref{app:engine-extended-design}.

\subsection{Timing Model and Data Pipeline}
\label{sec:timing-model}
\label{sec:data-engine}
\label{sec:learning}

The Timing Model (StreamingJev) turns the continuous observation stream into a decision
about whether the system should respond. Given causal observations $\mathcal I_{\leq t}$, standing constraints or
role context $c_t$, and output state $s_t$, the Timing Model predicts
\begin{equation}
    y_t = f_{\theta}(\mathcal I_{\leq t}, c_t, s_t),
    \qquad
    y_t \in \{\texttt{<silence>},\texttt{<response>}\}.
    \label{eq:engine-timing-decision}
\end{equation}
This decision controls the response channel while leaving content and
action planning to the Response Model. When output is inactive, \texttt{<response>} initiates a response
and \texttt{<silence>} maintains observation; when active, they indicate
continuation and release, respectively. Constraints may specify a
condition to monitor or broader responsibilities without naming each
emerging task.

Ordinary video descriptions identify events but rarely specify when
intervention becomes useful. Recent work therefore studies streaming speech and proactive dialogue as supervision for online assistants~\citep{chen2025livecc,zhang2025proassist}. Our {event-driven data engine}
combines Internet videos with existing video annotations to construct
this supervision. As illustrated in Figure~\ref{fig:event-data-engine},
it organizes actors, objects, and state changes into event-centered
records, associates them with relevant constraints, and identifies
response windows from supporting evidence. Full-video outcomes help
offline annotators locate precursor cues, but each training input
contains only observations available up to its decision time, together
with prior interaction state. A response window may precede the visible
outcome; an event boundary alone does not establish a need to intervene.

\begingroup
\colorlet{blue}{black}
\begin{figure}[!t]
    \centering
    \includegraphics[width=\linewidth]{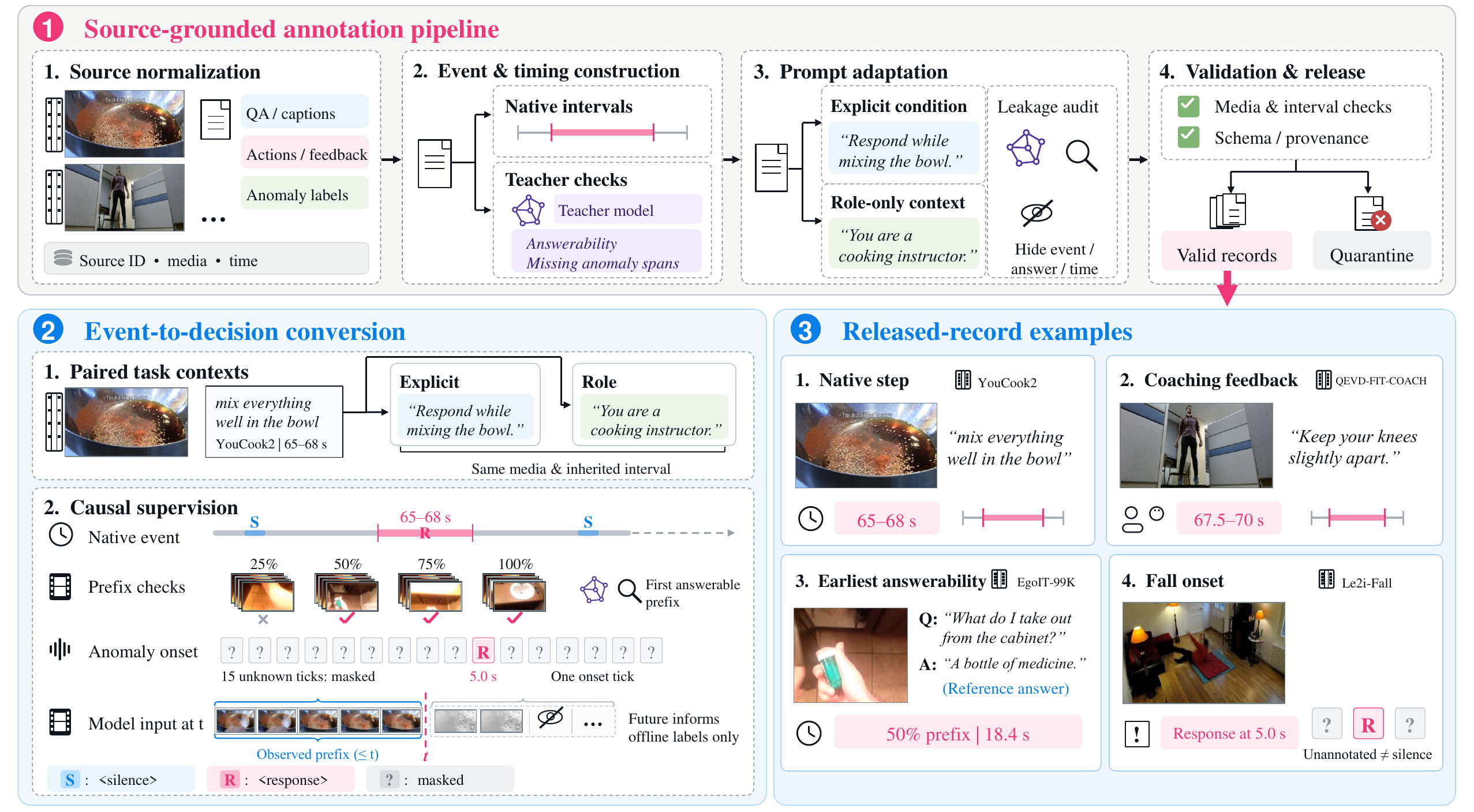}
    {\color{blue}
    \caption{\textbf{Event-driven supervision with causal model inputs.} Offline annotators use full videos to identify event context and intervention windows, while model inputs at time $t$ contain only observations up to $t$. Event descriptions ground the construction of explicit-condition and role- or anomaly-conditioned timing supervision; the Timing Model learns causal decision sequences, with unknown intervals excluded from supervision. 
    % Examples use released annotations and shortened English prompts.
    }
    \label{fig:event-data-engine}}
\end{figure}

\endgroup

Constructing these sequences requires both grounding an event in the video
and deciding when it becomes relevant to the system's responsibilities.
We use two forms of timing supervision. \emph{Explicit-condition timing}
identifies when evidence supports responding under a specified condition.
\emph{Open timing} uses role constraints or anomalous events to supervise
responses without naming the target event in advance. Event descriptions
and source annotations ground the construction of these labels.
The resulting sequences cover response-window entry, continuation, and
exit. Confirmed non-intervention periods provide silence labels,
including all-silence examples; unknown intervals are excluded from
supervision. Each training input combines causal visual observations
with task or role context and the current output state.

We post-train the Timing Model with a weighted autoregressive loss over
the supervised tokens in all decision turns:
\begin{equation}
    \mathcal L_{\mathrm{timing}}
    = -\frac{\sum_{\tau\in\mathcal B}\sum_j m_{\tau,j}w_{\tau,j}
       \log p_{\theta}(x_{\tau,j}\mid h_{\tau,j})}
       {\sum_{\tau\in\mathcal B}\sum_j m_{\tau,j}},
    \label{eq:engine-timing-loss}
\end{equation}
where $\mathcal B$ is the global batch, $x_{\tau,j}$ is a target token,
$h_{\tau,j}$ contains the causal input and preceding tokens, and
$m_{\tau,j}$ marks valid supervision. The weights $w_{\tau,j}$ depend
on the decision state; the denominator counts valid tokens.
We construct 224,268 training sequences from 20,258 source videos,
containing 2,058,267 supervised decision turns. StreamingJev is
post-trained from Qwen3.5-4B for three epochs, updating its language
model and output head while freezing the visual encoder and merger.

During training, each sequence starts from an idle output state, and
preceding ground-truth decisions update the state for subsequent observations.
Thus, supervision connects response initiation, continuation, and termination
across turns. Appendix~\ref{app:engine-learning-design} details sequence
construction and optimization.

\subsection{Efficient Streaming Inference}
\label{sec:timing-efficiency}

The Timing Model is evaluated repeatedly as observations arrive, making
the cost of each update central to continuous operation. We address this
cost at three stages of inference: processing the accumulated history,
representing newly observed visual changes, and generating a timing decision
(Figure~\ref{fig:streaming-efficiency}).

\begingroup
\colorlet{blue}{black}
\begin{figure}[!t]
    \centering
    \includegraphics[width=\linewidth]{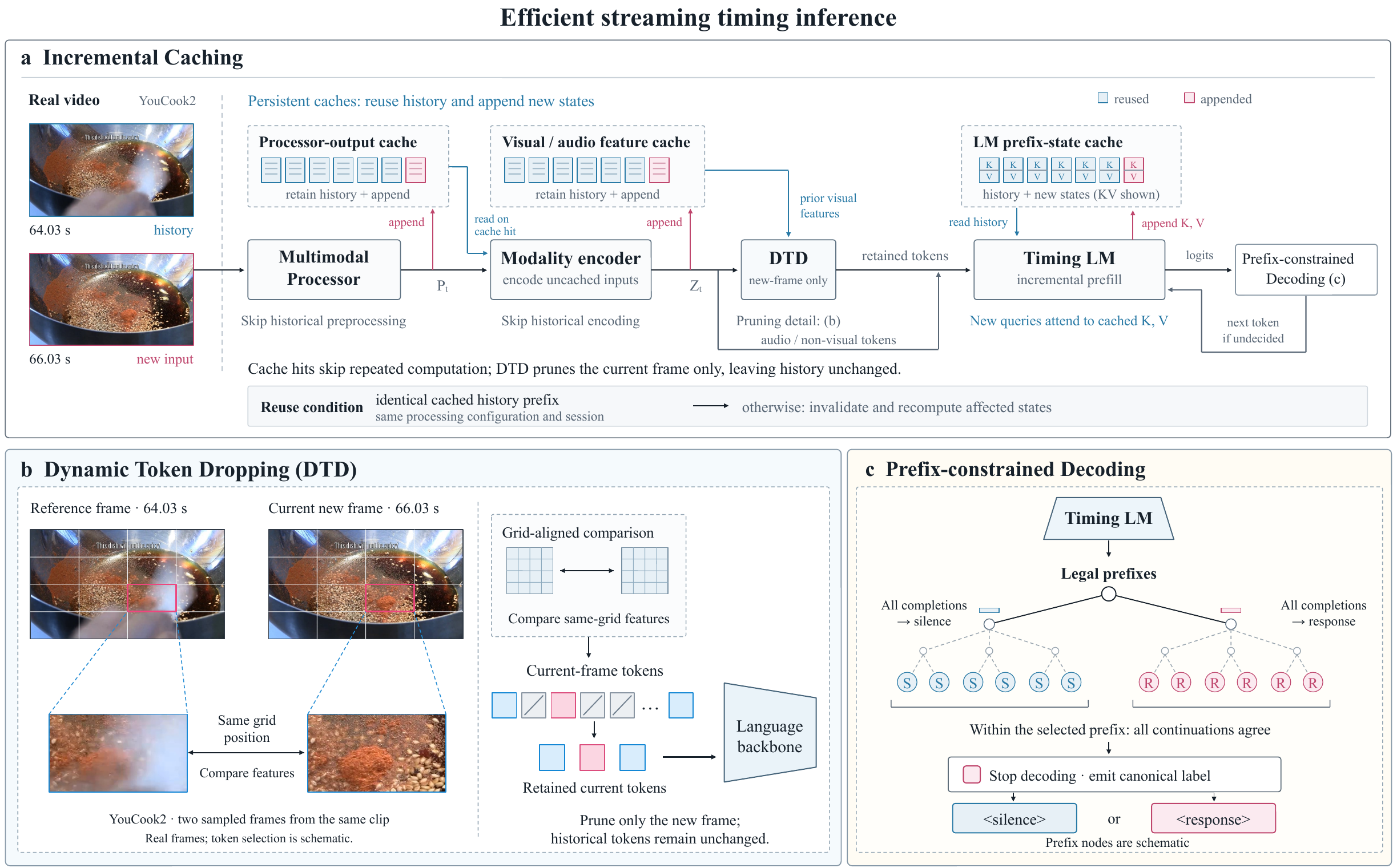}
    {\color{blue}
    \caption{\textbf{Three complementary efficiency mechanisms.} Incremental Caching reuses historical post-vision-encoder features and KV states subject to prefix and session validity. Dynamic Token Dropping (DTD) prunes redundant current-frame visual tokens using same-grid cross-frame similarity. Prefix-constrained Decoding terminates once all legal continuations imply the same binary decision and emits its canonical label. Token counts and the prefix tree are schematic; token counts do not imply a fixed retention ratio.}
    \label{fig:streaming-efficiency}}
\end{figure}
\endgroup

With \emph{Incremental Caching}, we cache historical post-vision-encoder features
and language-model key--value (KV) states. At each update, only newly
appended frames undergo visual preprocessing and encoding, while the
language model prefills only the uncached suffix. Cached states are isolated
by session and reused only when the input prefix and inference configuration
remain compatible.

Historical reuse does not eliminate redundancy within newly arriving
observations. Consecutive video frames often contain similar visual
features, even though only a small part of the scene has changed.
TimeChat-Online similarly exploits cross-frame redundancy through differential token dropping~\citep{yao2025timechatonline}.
\emph{Dynamic Token Dropping (DTD)} compares the current frame with the
preceding frame when their visual grids match. Specifically, it compares
features at the same token-grid locations. A current-frame token is
pruned as redundant when its cosine similarity exceeds the threshold
$\tau$, before it enters the language backbone. This reduces the sequence length processed by the language model and the growth of its KV cache.
Because intervention can depend on subtle visual changes, the resulting
computational savings must be considered together with decision agreement
and timing quality.

The restricted output space offers a further opportunity to reduce
inference overhead. Since the Timing Model returns one of two predefined
decisions, \emph{Prefix-constrained Decoding} restricts generation to valid decision
prefixes and terminates as soon as all legal continuations imply the
same decision. The runtime then emits the corresponding canonical label.
Together, Incremental Caching, Dynamic Token Dropping, and
Prefix-constrained Decoding
reduce work on historical context, incoming visual representations, and
output generation without changing the binary decision interface.
% Section~\ref{sec:efficiency-evaluation-draft} evaluates their cumulative
% effects on decision latency and memory usage.

\section{Experiments}
\label{sec:experiments-draft}

We evaluate intervention timing, delivered assistance, physical task completion,
and streaming efficiency. Detailed datasets, system configurations, and scoring
protocols are provided in Appendices~\ref{app:timing-protocols}--\ref{app:efficiency-protocols}.

\subsection{Proactive Timing Evaluation}
\label{sec:timing-evaluation-draft}

We compare timing on Omni-MMI PA/PT~\citep{ppb_omni2025} and paired
ProactiveVideoQA-derived episodes from EGO and WEB videos~\citep{wang2025proactivevideoqa}. \emph{Explicit}
specifies the assistance objective; \emph{Role} supplies standing responsibilities
without revealing the target event. Both use the same videos and references,
with reference annotations hidden during inference.
ProAgent receives observations every 200\,ms. The Base/Ours pair shares a
Qwen3.8-27B Worker and input and scheduling budgets; other systems retain their
respective interfaces and schedules. Responses are matched one-to-one to
reference windows using first visible-text times; unmatched responses are
counted against the reference annotations.

\begin{table}[!ht]
\centering\small
\caption{\textbf{Proactive timing performance.} PA/PT are evaluated on Omni-MMI; Explicit and Role use paired ProactiveVideoQA-derived episodes. PA, PT, and episode-averaged F1 are percentages. FP/min counts responses unmatched to reference windows per minute. A dash indicates an incompatible decision interface.}
\label{tab:timing-high-frequency}
\setlength{\tabcolsep}{4pt}
\renewcommand{\arraystretch}{0.95}
\setlength{\aboverulesep}{0.3ex}
\setlength{\belowrulesep}{0.4ex}
\begin{adjustbox}{max width=\linewidth}
\begin{tabular}{@{}lrrrrrr@{}}
\toprule
& \multicolumn{2}{c}{Omni-MMI} & \multicolumn{2}{c}{Explicit} & \multicolumn{2}{c}{Role} \\
\cmidrule(lr){2-3}\cmidrule(lr){4-5}\cmidrule(l){6-7}
System & PA $\uparrow$ & PT $\uparrow$ & F1 $\uparrow$ & FP/min $\downarrow$ & F1 $\uparrow$ & FP/min $\downarrow$ \\
\midrule
\multicolumn{7}{@{}l}{\textit{General-purpose: open weights}}\\
Qwen3.5-4B & 37.00 & 35.50 & 36.05 & 8.48 & 24.79 & 3.76 \\
Qwen3-VL-8B-Instruct & 53.00 & 77.00 & 33.37 & 11.81 & 23.94 & 2.67 \\
MiniCPM-V-4.6 & 23.50 & 81.00 & 28.25 & 20.06 & 28.74 & 19.48 \\
\midrule
\multicolumn{7}{@{}l}{\textit{General-purpose: closed source}}\\
GPT-6 & 29.00 & 44.00 & 32.44 & 2.49 & 25.66 & 1.32 \\
Gemini & 27.00 & 54.00 & 31.56 & 2.29 & 22.59 & 0.82 \\
Claude & 43.00 & 17.50 & 36.16 & 2.05 & 27.91 & 1.53 \\
\midrule
\multicolumn{7}{@{}l}{\textit{Streaming/proactive methods}}\\
LiveStar-8B & --- & --- & 34.16 & 9.71 & 31.98 & 7.65 \\
Dispider & 18.00 & 74.00 & 10.16 & 0.90 & 1.09 & \textbf{0.05} \\
\midrule
\multicolumn{7}{@{}l}{\textit{Matched Timing Models + shared Worker}}\\
Qwen3.5-4B Base + Worker & 35.50 & 53.00 & 30.49 & 0.70 & 28.72 & 0.62 \\
StreamingJev + Worker (Ours) & \textbf{57.00} & \textbf{87.00} & \textbf{42.43} & \textbf{0.04} & \textbf{33.45} & 0.18 \\
\bottomrule
\end{tabular}
\end{adjustbox}
\end{table}

Table~\ref{tab:timing-high-frequency} shows that ProAgent achieves the highest
PA, PT, and F1 scores. Explicit F1 reaches 42.43\%, exceeding the strongest
baseline by 6.27 percentage points, with only 0.04 unmatched responses per
minute. Under Role constraints, ProAgent leads LiveStar by 1.47 points while
reducing FP/min from 7.65 to 0.18. Dispider produces fewer unmatched responses
in this condition but attains only 1.09\% F1, illustrating why restraint must
be considered together with intervention coverage.

With the Worker fixed, training the Timing Model improves Explicit and Role
F1 by 11.94 and 4.73 points, respectively. Role F1 remains below Explicit F1,
indicating that identifying an intervention from broad responsibilities is
more challenging than responding to a specified objective.

\subsection{Proactive Interaction Evaluation}
\label{sec:interaction-evaluation-draft}

A well-timed response is useful only if it conveys the information needed
in that situation. Each system responds at its own trigger times, without
reference triggers or a second answer-generation pass. We assess these
same trajectories using \emph{Outcome}, which
measures complete reference-answer coverage by responses initiated within
assistance windows,
and ProactiveVideoQA's \emph{PAUC}~\citep{wang2025proactivevideoqa}, which
jointly rewards correctness and timeliness ($\omega=0.5$). Coverage is scored
per reference target and averaged within and then across episodes.

ProAgent achieves the highest PAUC under both conditions: 0.6131 for Explicit
and 0.5121 for Role (Table~\ref{tab:role-high-frequency-outcome}). Under Role,
it also leads in Outcome. Under Explicit, GPT-6 provides higher complete-answer
coverage (0.1523 versus 0.1432), while ProAgent leads in PAUC. The two measures
capture complementary properties: PAUC credits timely partial information,
whereas Outcome requires complete coverage. Thus, strong answer generation
alone does not ensure the best time-aware assistance.

The three closed-source systems incur mean first-text latencies of
14.55--19.81\,s, compared with 1.07--1.14\,s for ProAgent. Their remote response
path includes network transfer and service processing; extended chain-of-thought
reasoning, when enabled, can further delay visible output. Such delays can
move responses beyond useful assistance windows, limiting time-sensitive
performance. ProAgent separates frequent timing decisions from asynchronous
response generation (Figure~\ref{fig:propa-engine}), combining short response latency with the highest
PAUC in the evaluated configurations.

\begin{table}[!t]
\centering\small
\caption{\textbf{Helpfulness and runtime performance of self-triggered responses.} Outcome and PAUC are reported on a 0--1 scale. Latency is the mean time to first visible response text (s). Runtime protocols are specified in Appendix~\ref{app:condition-runtime}.}
\label{tab:role-high-frequency-outcome}
\setlength{\tabcolsep}{5pt}
\renewcommand{\arraystretch}{0.95}
\setlength{\aboverulesep}{0.3ex}
\setlength{\belowrulesep}{0.4ex}
\begin{adjustbox}{max width=\linewidth}
\begin{tabular}{@{}lrrrrrr@{}}
\toprule
& \multicolumn{3}{c}{Explicit} & \multicolumn{3}{c}{Role} \\
\cmidrule(lr){2-4}\cmidrule(l){5-7}
System & Latency (s) $\downarrow$ & Outcome $\uparrow$ & PAUC $\uparrow$ & Latency (s) $\downarrow$ & Outcome $\uparrow$ & PAUC $\uparrow$ \\
\midrule
\multicolumn{7}{@{}l}{\textit{General-purpose: open weights}}\\
Qwen3.5-4B & 2.09 & 0.1263 & 0.4086 & 2.05 & 0.0723 & 0.3173 \\
Qwen3-VL-8B-Instruct & 1.55 & 0.1369 & 0.4633 & 1.42 & 0.0462 & 0.3063 \\
MiniCPM-V-4.6 & 1.22 & 0.0965 & 0.5396 & 1.21 & 0.0580 & 0.3747 \\
\midrule
\multicolumn{7}{@{}l}{\textit{General-purpose: closed source}}\\
GPT-6 & 16.59 & \textbf{0.1523} & 0.4123 & 19.81 & 0.0336 & 0.2603 \\
Gemini & 14.58 & 0.0239 & 0.3379 & 14.96 & 0.0201 & 0.2546 \\
Claude & 14.55 & 0.0885 & 0.3645 & 14.55 & 0.0431 & 0.2964 \\
\midrule
\multicolumn{7}{@{}l}{\textit{Streaming/proactive methods}}\\
LiveStar-8B & 2.53 & 0.0372 & 0.3185 & 2.31 & 0.0481 & 0.2859 \\
Dispider & 1.62 & 0.0057 & 0.2645 & 2.85 & 0.0000 & 0.2382 \\
\midrule
\multicolumn{7}{@{}l}{\textit{Matched Timing Models + shared Worker}}\\
Qwen3.5-4B Base + Worker & 1.09 & 0.0826 & 0.3355 & 1.16 & 0.0491 & 0.2828 \\
StreamingJev + Worker (Ours) & \textbf{1.07} & 0.1432 & \textbf{0.6131} & \textbf{1.14} & \textbf{0.1231} & \textbf{0.5121} \\
\bottomrule
\end{tabular}
\end{adjustbox}
\end{table}

\begin{figure}[!t]
    \centering
    \includegraphics[width=0.84\linewidth]{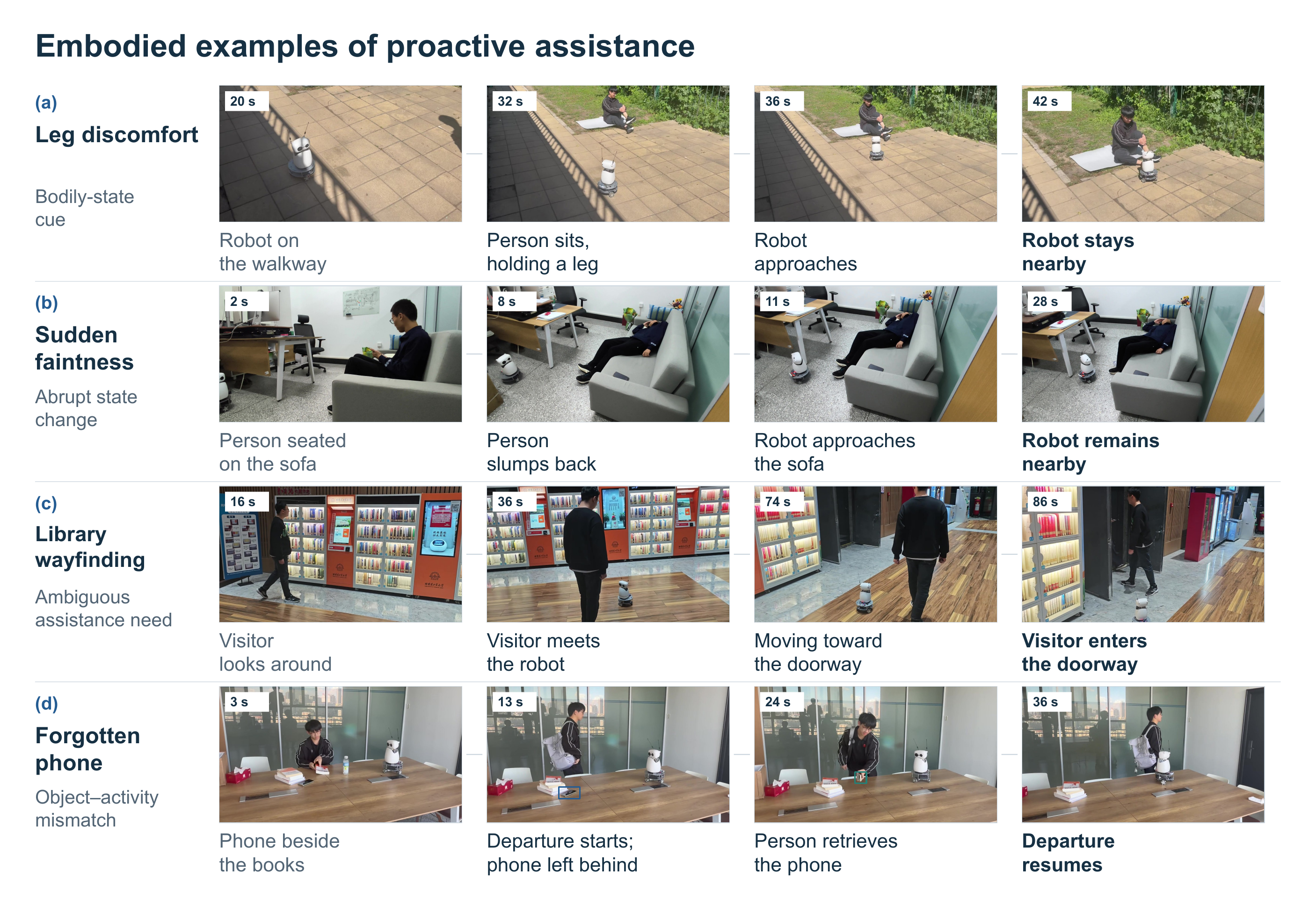}
    \caption{\textbf{Recorded embodied interaction trajectories.} Four frames per row show changes in the scene, robot interaction, and subsequent observations. (a) A person sits down holding their lower leg; the robot approaches and stays nearby. (b) A person slumps back onto a sofa; the robot approaches and remains nearby. (c) A visitor encounters the robot, moves with it toward a doorway, and enters. (d) A person preparing to leave returns to retrieve a phone before resuming departure. External-camera recordings document visible behavior; frame labels describe the observed actions and states. Timestamps are video positions, not model-trigger times. Quantitative performance is measured in the separate 60-case evaluation (Table~\ref{tab:physical-main-draft}).}
    \label{fig:embodied-case-trajectories}
\end{figure}

\subsection{Embodied Evaluation}
\label{sec:physical-evaluation-draft}

We evaluate 60 Reachy Mini cases, with 20 each for Guidance, Warning, and
Interaction. These categories cover assistance during ongoing activities,
responses to emerging errors or risks, and proactive engagement with human
needs, respectively. Systems receive standing responsibilities and available
capabilities, without reference triggers or required responses. \emph{Time}
measures mean absolute intervention-time error; \emph{Outcome} measures
completion of required events with correct content and entities, assigning
zero when prescribed ordering constraints are violated.

ProAgent attains the highest Outcome in all three categories
(Table~\ref{tab:physical-main-draft}), exceeding the strongest baseline in each
by 0.089, 0.066, and 0.173, respectively. The largest completion-score gain
occurs in Interaction, where needs must be recognized through proactive
engagement. These results extend the evidence beyond verbal assistance:
the system connects inferred needs to completed events involving people
and available device capabilities.

\begin{table}[!ht]
\caption{\textbf{Embodied evaluation.} Each category contains 20 cases. Time: mean absolute timing error (s); Outcome: completion score on a 0--1 scale.}
\label{tab:physical-main-draft}
\centering\small
\setlength{\tabcolsep}{6pt}
\renewcommand{\arraystretch}{0.95}
\setlength{\aboverulesep}{0.3ex}
\setlength{\belowrulesep}{0.4ex}
\begin{adjustbox}{max width=\linewidth}
\begin{tabular}{@{}lrrrrrr@{}}
\toprule
& \multicolumn{2}{c}{Guidance}
& \multicolumn{2}{c}{Warning}
& \multicolumn{2}{c}{Interaction} \\
\cmidrule(lr){2-3}\cmidrule(lr){4-5}\cmidrule(l){6-7}
Method & Time $\downarrow$ & Outcome $\uparrow$
& Time $\downarrow$ & Outcome $\uparrow$
& Time $\downarrow$ & Outcome $\uparrow$ \\
\midrule
GPT-6-Astra & 12.56 & 0.759 & 4.08 & 0.712 & 11.06 & 0.676 \\
Gemini 3 & 5.50 & 0.821 & 1.62 & 0.827 & 9.15 & 0.640 \\
Claude Fable 5 & 9.29 & 0.480 & 21.39 & 0.326 & 10.09 & 0.189 \\
Qwen3.5VL 27B & 11.21 & 0.748 & 2.35 & 0.719 & 9.03 & 0.618 \\
ProAgent & \textbf{1.28} & \textbf{0.910} & \textbf{1.26} & \textbf{0.893} & \textbf{1.98} & \textbf{0.849} \\
\bottomrule
\end{tabular}
\end{adjustbox}
\end{table}

The recorded trajectories in Figure~\ref{fig:embodied-case-trajectories} show four interactions, linking emerging
assistance needs to the observed behavior that follows. These cases also highlight people as possible executors of physical tasks: in the phone-retrieval scenario, the person retrieves the object rather than the robot manipulating it. Appendix~\ref{app:embodied-case-studies} relates these trajectories to intervention decisions and execution feedback.

\Needspace{19\baselineskip}
\subsection{Analysis and Efficiency}
\label{sec:efficiency-evaluation-draft}

\begingroup
\setlength{\intextsep}{0pt}
\begin{wraptable}{r}{0.55\linewidth}
\setlength{\abovecaptionskip}{0pt}
\setlength{\belowcaptionskip}{6pt}
\caption{\textbf{Cumulative inference optimizations.} Latency excludes image decoding and preprocessing (ms); Mem. is peak allocated GPU memory (GB); Agree. is decision agreement (\%) with full recomputation. DTD uses $\tau=0.9$.}
\label{tab:efficiency-main-draft}
\centering\small
\setlength{\tabcolsep}{3pt}
\renewcommand{\arraystretch}{1.10}
\setlength{\aboverulesep}{0.4ex}
\setlength{\belowrulesep}{0.65ex}
\begin{adjustbox}{max width=\linewidth}
\begin{tabular}{@{}lrrrr@{}}
\toprule
Configuration & P50 $\downarrow$ & P95 $\downarrow$ & Mem. $\downarrow$ & Agree. $\uparrow$ \\
\midrule
Recompute & 346.84 & 512.63 & 9.85 & 100.00 \\
+ Cache & 240.26 & 270.06 & 9.63 & 100.00 \\
+ DTD & 214.16 & 269.73 & 9.50 & 98.65 \\
+ Constrained & 147.05 & 251.77 & 9.50 & 98.65 \\
\bottomrule
\end{tabular}
\end{adjustbox}
\end{wraptable}
We evaluate cumulative inference optimizations on 398 causal video
sequences comprising 3,768 timing decisions, using identical inputs and
frozen interaction histories across configurations. Frames are sampled
at 1\,Hz, and all configurations run sequentially on a single NVIDIA
A800 80\,GB GPU with BF16 precision.
Table~\ref{tab:efficiency-main-draft} measures the cumulative effects of
incremental caching, dynamic token dropping, and prefix-constrained decoding on the trained
Timing Model.

The combined configuration reduces median latency from 346.84\,ms to
147.05\,ms and P95 latency from 512.63\,ms to 251.77\,ms, with 98.65\%
decision agreement relative to full recomputation. Latency measures
synchronized inference time from preprocessed input to decision
completion, excluding image decoding and preprocessing. Decision
agreement measures the effect on model outputs, not task accuracy.
The measurement protocol and component analysis are detailed in
Appendix~\ref{app:efficiency-protocols}.
\par
\endgroup

\section{Discussion}
\label{sec:discussion}

The matched-Worker comparison highlights the importance of intervention
quality: first-text latency changes little, yet PAUC rises
from 0.3355 to 0.6131 under Explicit and from 0.2828 to 0.5121 under Role.
These gains support learned intervention decisions beyond
faster response generation alone.

ProPAI shifts the source of an action from an explicit task to a need
identified under persistent responsibilities. ProAgent provides a modular
realization of this shift, while leaving important distinctions for
evaluation: recognizing a supplied condition is not unrestricted task
discovery, and a timely response in recorded video does not establish
physical benefit. Broader evaluation must test whether interventions
resolve needs across changing roles and capabilities, including when
silence or clarification is preferable. Permission to act remains
distinct from an inferred need.

These questions connect future improvements in memory, learning, and
architecture. Memory should retain unresolved commitments while revising
stale beliefs; learning should distinguish intervention value from event
timing, executor failure, and recovery caused by others. Native models
could jointly represent timing and response, with adaptive computation
that preserves observation continuity and response deadlines. Callable
skills, including specialized action policies, could broaden feasible
interventions while retaining explicit authorization and human control.
Appendices~\ref{app:engine-extended-design} and~\ref{app:engine-open-challenges}
develop these directions and their evaluation requirements.

\Needspace{12\baselineskip}
\section{Conclusion}
\label{sec:conclusion}

We introduce \textbf{Proactive Physical AI} and \textbf{ProAgent}, a system
that connects continuous observation to timely communication and physical
action under standing responsibilities. Causal video supervision makes
intervention timing learnable, while the asynchronous separation of timing
and response supports ongoing monitoring during generation and execution.
Experiments show stronger intervention timing and time-aware assistance
than the evaluated alternatives, together with improved event completion
across three embodied task categories. Streaming optimizations reduce the
cost of repeated decisions while maintaining high decision agreement.
Together, these findings support a practical approach to physical AI that
identifies emerging needs and coordinates responses through available
human and device capabilities.

\FloatBarrier

\endgroup
\bibliography{iclr2027_conference}
\bibliographystyle{iclr2027_conference}
\clearpage
\appendix

\numberwithin{table}{section}
\numberwithin{figure}{section}
\renewcommand{\theHtable}{appendix.\thesection.\arabic{table}}
\renewcommand{\theHfigure}{appendix.\thesection.\arabic{figure}}
\numberwithin{equation}{section}

\section{Connections to existing studies}
\label{app:connections}
\label{sec:related}

ProPAI draws on established research in streaming multimodal interaction, proactive robotics, embodied execution, and learning from observation. We organize these foundations around how an agent continuously observes the world, recognizes a need for assistance, and turns that need into an effective intervention.

\paragraph{Streaming Multimodal Models and Proactive Assistants.}
Streaming models make continuously arriving observations available for online reasoning. VideoLLM-online develops causal streaming dialogue training and efficient inference, while Dispider separates continuous perception, response decisions, and asynchronous generation~\citep{chen2024videollmonline,qian2025dispider}. Proactive interaction is also an explicit learning objective: MMDuet2 trains models to choose between responding and remaining silent across multiple turns, and ProactiveVideoQA evaluates response timing together with answer quality~\citep{wang2025mmduet2,wang2025proactivevideoqa}. StreamChat updates visual context during generation, while LiveStar integrates streaming perception, silence decisions, and memory management~\citep{liu2024streamchat,yang2025livestar}. Related approaches reduce streaming computation through event-gated processing and visual token pruning~\citep{ding2025streammind,yao2025timechatonline,zhang2026querystream}. ViSpeak studies responses to visual instructions, and Eyes Wide Open studies proactive responses in egocentric streams~\citep{fu2025vispeak,zhang2025eyeswideopen}. These advances provide both a computational foundation and evaluation tools for persistent assistants. Answering a specified question at the right moment is one important setting; identifying which need deserves a response under broader responsibilities introduces an additional decision about the task itself.

\paragraph{Proactive Human--Robot Interaction and Assistance.}
Robot initiative has a substantial research history. Existing reviews define proactive human--robot interaction and organize its varied behaviors and technical approaches~\citep{vandenbroek2024proactivehri}. Recent systems connect these ideas to foundation models: RoboOmni combines multimodal intent recognition, interaction confirmation, and manipulation~\citep{wang2025roboomni}; ProAct couples long-horizon social reasoning with low-latency multimodal behavior on a humanoid robot~\citep{zhang2026proact}; and event-driven assistive manipulation infers goals and intervention decisions from human-induced workspace state changes~\citep{liu2026eventdriven}. These are direct foundations for ProPAI, including precedents for assistance without an explicit task instruction. They also reveal distinct sources of initiative: inferred human intent, social context, and changes in the environment. Code-as-Monitor uses visual programs to monitor constraints and anticipate robotic failures~\citep{zhou2025codeasmonitor}. Beyond robotics, sensory-context assistants and mobile agents study how contextual observations reveal needs and support proactive assistance~\citep{yang2025proagent,kong2026proactivemobile}. ProPAI brings these sources into a shared formulation in which persistent roles and constraints guide need discovery, while available people and devices determine feasible responses.

\paragraph{Physical Reasoning and Capability-Grounded Execution.}
Vision-language-action models such as RT-2 and $\pi_{0.5}$ connect semantic knowledge to robot control and improve generalization across objects, instructions, and environments~\citep{brohan2023rt2,physicalintelligence2025pi05}. Complementary approaches expose reusable skills to higher-level reasoning: SayCan grounds language-based planning in skill affordances, Code as Policies composes perception and control APIs, and Inner Monologue incorporates environmental feedback into planning~\citep{ahn2022saycan,liang2022codeaspolicies,huang2022innermonologue}. VoxPoser grounds language in three-dimensional value maps for manipulation, while OK-Robot integrates visual-language perception, navigation, and grasping~\citep{huang2023voxposer,liu2024okrobot}. These methods provide execution and feedback mechanisms that proactive systems can build upon. Our function-call interface follows this programmable-capability perspective, exposing registered functions without prescribing their underlying controllers. The additional proactive question is how available capabilities affect whether, when, and through whom an emerging need should be addressed.

\paragraph{Learning from Human Video and Interaction.}
Human activity and assistance provide supervision beyond robot trajectories. HoloAssist captures collaborative task performance and supports mistake detection and intervention prediction, while RoboVQA combines human and robot experience for grounded long-horizon reasoning~\citep{wang2023holoassist,sermanet2023robovqa}. Predictive video learning offers another foundation: V-JEPA~2 combines large-scale video pretraining with robot interaction data to support action-conditioned prediction and planning~\citep{assran2025vjepa2}. LiveCC and ProAssist further connect video observations to streaming language and proactive assistant dialogue~\citep{chen2025livecc,zhang2025proassist}. These results motivate learning broadly transferable knowledge from observation while retaining embodiment-specific execution. For proactive learning, observing what happens next must additionally be connected to whose responsibility it is to intervene and when intervention is justified; this motivates our timing-supervision pipeline and the need for deployment feedback.

Taken together, these foundations motivate a coupled research problem rather than independent perception, timing, and control tasks. The same world event may call for silence under one role and assistance under another; choosing between a nearby speaker and a mobile robot changes the required start time; and successful intervention changes what should happen next. ProPAI organizes these dependencies into a continuous observation--intervention--feedback loop, providing a common setting for studying task discovery, timely restraint, and capability-grounded action.

\section{Timing Evaluation Protocols and Supplementary Results}
\label{app:timing-protocols}

This appendix evaluates intervention timing under benchmark-provided
constraints. It complements the real-time, role-conditioned evaluation
in Appendix~\ref{app:role-proactive-protocol}: here the relevant question
or condition is given, and the score measures readiness or first-trigger
timing rather than unrestricted need discovery. No real-time processing
deadline is imposed.

\subsection{Benchmark Tasks and Coverage}

Table~\ref{tab:timing-coverage} summarizes the selected tasks. OVO-Bench
supplies independent observation checkpoints, whereas StreamingBench
requires a first response during sequential observations.
StreamingBench is distinct from the streaming question-answering
benchmark StreamBench v0.3.

\begin{table}[htbp]
\centering\small
\caption{\textbf{Tasks used for timing evaluation.} Counts are scoring units in the complete benchmark releases.}
\label{tab:timing-coverage}
\setlength{\tabcolsep}{3pt}
\begin{tabular}{@{}p{0.15\linewidth}p{0.25\linewidth}p{0.28\linewidth}p{0.15\linewidth}r@{}}
\toprule
Benchmark / task & Input and constraint & Required decision & Metric & Full $N$ \\
\midrule
OVO / SSR & Video clip and specified procedural step &
Whether that step is currently being performed & Accuracy & 629 \\
OVO / CRR & Video clip and question &
Whether the available evidence is sufficient to answer & Accuracy & 240 \\
StreamingBench / PO & Sequential video observations and response condition &
When to trigger the first response & First-trigger accuracy & 250 \\
\bottomrule
\end{tabular}
\end{table}

The full OVO selection contains 869 timing probes; its content-answering
tasks are not additional timing examples. The 250 PO questions span 50
videos. All evaluated systems use these complete task releases.
The source-reported Dispider PO result is identified separately.

\subsection{Public-Benchmark Timing Comparisons}
\label{app:timing-public-comparison}

\begin{table}[!ht]
\centering\small
\caption{\textbf{Intervention timing under benchmark-provided constraints.} OVO-Bench measures response readiness at given checkpoints; StreamingBench measures first-response timing over sequential observations. Scores are percentages; higher is better. $\dagger$ denotes semantic output mapping. $*$ marks a result reported in the original Dispider paper~\citep{qian2025dispider}.}
\label{tab:timing-main-draft}
\setlength{\tabcolsep}{6pt}
\begin{tabular}{@{}l@{\hspace{1em}}rrr@{}}
\toprule
& \multicolumn{2}{c}{OVO-Bench} & StreamingBench \\
\cmidrule(lr){2-3}\cmidrule(l){4-4}
Timing Model & SSR & CRR & PO \\
\midrule
\multicolumn{4}{@{}l}{\textit{Closed-source general-purpose models}}\\
GPT-6 & 73.93 & 66.25 & 73.20 \\
Gemini 3.7 Flash & 78.22 & 84.17 & 80.00 \\
Claude Fable 5 & 82.67 & 74.58 & 71.60 \\
\midrule
\multicolumn{4}{@{}l}{\textit{Open-weight general-purpose models}}\\
Qwen3-VL-8B-Instruct & 64.39 & 43.33 & 49.20 \\
MiniCPM-V-4.6 & 31.96 & 50.00 & 19.60 \\
LLaVA-OneVision-7B & 65.50 & 57.92 & 28.80 \\
\midrule
\multicolumn{4}{@{}l}{\textit{Streaming and proactive methods}}\\
LiveStar-8B & 6.04 & 3.75 & 13.60 \\
Dispider & 37.52 & 48.75 & $25.34^{*}$ \\
\midrule
\multicolumn{4}{@{}l}{\textit{Source backbone and trained Timing Model}}\\
Qwen3.5-4B (Base)$^{\dagger}$ & 52.15 & 43.75 & 11.60 \\
StreamingJev (Ours) & 75.52 & 83.33 & 57.20 \\
$\Delta$ vs. Base (pp) & +23.37 & +39.58 & +45.60 \\
\bottomrule
\end{tabular}
\end{table}

The source-backbone comparison fixes Qwen3.5-4B and compares its unmodified
checkpoint with \textbf{StreamingJev}, post-trained for three epochs.
The trained model improves SSR from 52.15\% to 75.52\%, CRR from 43.75\%
to 83.33\%, and PO from 11.60\% to 57.20\%, corresponding to gains of
23.37, 39.58, and 45.60 percentage points, respectively. These gains cover
readiness at supplied checkpoints and sequential first-response timing.
The table also includes general-purpose models and task-conditioned
evaluations of streaming methods. Evaluation counts and visual
preprocessing are specified in the protocol below. The Dispider PO
score is cited from its original paper; the other scores are measured
in our evaluation.

\subsection{Inputs, Decision Interfaces, and Scoring}

\paragraph{Decision interfaces.}
\label{app:timing-decoding}
StreamingJev predicts \texttt{<response>} or \texttt{<silence>}.
Its source-backbone comparison uses the same binary decision task,
without a Worker or answer-quality judge; the output channel remains
idle between decisions, so this comparison does not test interruption
of an ongoing reply. General-purpose response models use the
corresponding benchmark question-answering interface rather than being
required to emit the Timing Model's labels.

For the Base row marked $\dagger$, a fixed, label-independent decoder accepts
the canonical labels, maps \texttt{START}/\texttt{CONTINUE} to response
and \texttt{WAIT}/\texttt{STOP}/\texttt{YIELD} to silence, and accepts
the bare strings \texttt{response}/\texttt{silence}. A complete wrapper
such as \texttt{<response>WAIT</response>} is interpreted by its enclosed
action. Matching is case-sensitive after trimming outer whitespace;
ambiguous prose, incomplete wrappers, and conflicting actions remain
invalid. The mapping is applied to saved outputs without changing
generation histories and uses neither reference answers nor an LLM
judge. Canonical outputs from StreamingJev are unchanged.

\paragraph{OVO-Bench.}
\label{app:timing-ovo}
Each SSR or CRR probe is evaluated independently using its released
clip and question. SSR concerns the specified step, not completion
of the entire procedure; CRR concerns whether the question is answerable
from the available evidence. For the binary Timing Models, Yes and No
labels correspond to response and silence. Benchmark-interface VLM
answers are scored with the released answer parser. All probes,
including invalid predictions, remain in the denominator:
\begin{equation}
    \operatorname{Acc}_k
    = \frac{1}{N_k}\sum_{i=1}^{N_k}
      \mathbf{1}\{\widehat y_i=y_i\},
    \qquad k\in\{\mathrm{SSR},\mathrm{CRR}\}.
\end{equation}
Probe histories are not accumulated across clips. Closed-source
models use at most 64 uniformly sampled frames per released clip,
encoded as JPEG images with a maximum side length of 512 pixels.
MiniCPM-V-4.6 uses its video processor at 1 fps, capped at 128 frames
with uniform subsampling for longer clips. These are model-specific
input budgets, not a controlled equal-frame comparison.

\paragraph{StreamingBench.}
\label{app:timing-streaming}
For the Base and StreamingJev comparison, let $s_i$ be the annotated
query-start time and $g_i$ the reference response time. The adapter
evaluates at $s_i+1,s_i+2,\ldots,g_i+4$, stopping at the first response
under the selected decoder. Each input contains the current available
frame, the response condition and prescribed output string, the stream
time, and up to 30 seconds of observation/decision history. A frame is
eligible only if its presentation timestamp does not exceed the
decision time. Inputs are visual-only, with temperature zero and at
most eight generated tokens per decision. The reference time and
evaluation horizon are not supplied in the prompt.

Closed-source models follow the benchmark's once-per-second Yes/No
readiness query and generate an answer after the first Yes, using
causal 1 fps frame histories over the evaluation horizon. Response
timing is measured in source-video timestamps.

Let $\tau_i$ be the first response time, or $\varnothing$ if there is
no response within the evaluation horizon. Timing accuracy is
\begin{equation}
    A_{\mathrm{PO}}
    =\frac{1}{N}\sum_{i=1}^{N}
      \mathbf{1}\{\tau_i\neq\varnothing
      \ \land\ |\tau_i-g_i|\leq2\}.
\end{equation}
The boundary is inclusive. A response before $g_i-2$ is early,
one after $g_i+2$ is late, and no response is a miss; a later correct
trigger cannot repair an earlier first response.
Neither the timing-only score nor a supplied response condition
tests open-ended answer generation by StreamingJev.

\subsection{Timing Improvements and Error Analysis}
\label{app:timing-error-analysis}

Table~\ref{tab:timing-correct-counts} reports the correct counts underlying
the source-backbone comparison. StreamingJev correctly classifies
475 of 629 SSR probes and 200 of 240 CRR probes, compared with 328 and
105 for Base, respectively. On PO, it triggers within the reference
window on 143 of 250 questions, compared with 29 for Base.
These counts yield gains of 23.37, 39.58, and 45.60 percentage points.

\begin{table}[htbp]
\centering\small
\caption{\textbf{Correct counts for StreamingJev versus Base.}
Both models are evaluated on the same probes or questions for each task.
Differences are computed from the counts before rounding.}
\label{tab:timing-correct-counts}
\setlength{\tabcolsep}{6pt}
\begin{tabular}{@{}lrrrr@{}}
\toprule
Task & Total & Base correct & Ours correct & $\Delta$ (pp) \\
\midrule
OVO SSR & 629 & 328 & 475 & +23.37 \\
OVO CRR & 240 & 105 & 200 & +39.58 \\
StreamingBench PO & 250 & 29 & 143 & +45.60 \\
\bottomrule
\end{tabular}
\end{table}

\paragraph{First-response errors.}
Of the 250 PO questions, StreamingJev produces 143 on-time responses
(57.20\%), 71 early responses (28.40\%), 10 late responses (4.00\%),
and 26 misses (10.40\%). These mutually exclusive categories account
for all questions. Early responses constitute 71 of the 107 errors
(66.36\%), making premature intervention the largest error category.
The remaining errors comprise late responses and missed interventions;
thus, first-trigger accuracy distinguishes timely intervention from
merely producing an eventual response.

\FloatBarrier

\section{Response Capability Evaluation}
\label{app:interaction-protocols}

We evaluate answer generation on OVO-Bench, Omni-MMI, and StreamBench v0.3
to characterize the perception, temporal reasoning, and conversational
capabilities available to the Response Model. Each benchmark supplies the
question and answering opportunity; the model generates an answer rather
than deciding when to initiate an intervention. These measurements complement
the timing evaluation in Appendix~\ref{app:timing-protocols}, without
combining the two into a single score.

\subsection{Tasks and Evaluation Coverage}
\label{app:worker-supporting-capabilities}

\begin{table}[htbp]
\centering\small
\caption{Response-evaluation tasks and full-release coverage.}
\label{tab:interaction-coverage}
\setlength{\tabcolsep}{4pt}
\begin{tabular}{@{}p{0.19\linewidth}p{0.23\linewidth}p{0.51\linewidth}@{}}
\toprule
Benchmark & Evaluated units & Capability and metric \\
\midrule
OVO-Bench & 1,468 multiple-choice probes; 698 counting probes &
Visual perception, backward tracing, and repetition counting; task accuracy and within-category macro accuracy. \\
Omni-MMI & 800 records; 1,690 answers &
Action prediction (AP), state grounding (SG), and multi-turn dependency (MD); record accuracy, requiring all answers correct for SG/MD. \\
StreamBench v0.3 & 307 records; 1,838 questions &
Search, memory, and conversational answering; semantic accuracy per task and a four-task macro. \\
\bottomrule
\end{tabular}
\end{table}

Table~\ref{tab:interaction-content-main} summarizes the response-side
measurements. Models answer through their language-generation interfaces;
StreamingJev is not used as an answer generator. LiveStar and Dispider
are evaluated through their question-conditioned response paths, not their
autonomous triggering policies. OVO and Omni-MMI use visual
inputs without audio or ASR; StreamBench uses causal visual histories and
ASR text. Input budgets are model-specific.

\begin{table}[htbp]
\centering\small
\caption{Response capability at benchmark-supplied answering opportunities (\%). OVO columns are task macros; Omni SG/MD require all-step correctness; StreamBench reports the selected core-task macro. $S$ marks the selected-clip configuration described below; dashes indicate unavailable scores.}
\label{tab:interaction-content-main}
\setlength{\tabcolsep}{4pt}
\resizebox{\linewidth}{!}{%
\begin{tabular}{@{}lrrrrrr@{}}
\toprule
& \multicolumn{2}{c}{OVO-Bench} & \multicolumn{3}{c}{Omni-MMI} & StreamBench v0.3 \\
\cmidrule(lr){2-3}\cmidrule(lr){4-6}\cmidrule(l){7-7}
Response Model & Perception & Backward & AP & SG & MD & Core-task macro \\
\midrule
\multicolumn{7}{@{}l}{\textit{Open-weight general-purpose models}}\\
Qwen3-VL-8B-Instruct & 61.13 & 41.95 & 35.50 & 11.33 & 13.67 & 49.91 \\
MiniCPM-V-4.6 & 62.30 & 45.27 & 38.00 & 9.00 & 9.33 & 60.38 \\
LLaVA-OneVision-7B & 64.33 & 44.11 & 23.50 & 5.00 & 5.67 & 52.27 \\
Qwen3.5-4B (source backbone)$^{S}$ & 66.98 & 53.81 & 35.00 & 13.00 & 11.67 & 50.54 \\
\midrule
\multicolumn{7}{@{}l}{\textit{Streaming/proactive methods: response-path evaluation}}\\
LiveStar-8B & 58.32 & 43.70 & 30.00 & 14.33 & 4.00 & 45.00 \\
Dispider & 54.34 & 36.06 & 0.00 & 5.33 & 6.67 & --- \\
\midrule
\multicolumn{7}{@{}l}{\textit{ProAgent response backbone}}\\
Qwen3.8-27B$^{S}$ & 74.64 & 65.35 & 43.00 & 19.00 & 16.33 & 55.03 \\
\bottomrule
\end{tabular}%
}
\end{table}

\subsection{OVO-Bench: Perception and Temporal Evidence}
\label{app:interaction-ovo}

The six perception tasks are optical character recognition (OCR), action
recognition (ACR), attribute recognition (ATR), spatial understanding (STU),
future prediction (FPD), and object recognition (OJR). Backward tracing
comprises episodic memory (EPM), action sequence identification (ASI), and
hallucination detection (HLD). Repetition event count (REC) is reported
separately. The perception and backward summaries give their constituent
tasks equal weight, rather than pooling probes. SSR and CRR belong to the
timing evaluation and are not included in these averages.

Models receive the released task prompt and corresponding video input,
without reference answers. The released scorer checks multiple-choice
answer letters and extracts digit groups for REC; it does not semantically
rewrite model outputs. All probes remain in the denominator.
Qwen3-VL-8B uses the native video interface with 64 frames and a 4,096-token
output limit. LiveStar uses its 32-segment, 448-pixel visual preprocessing
and a 1,024-token output limit. MiniCPM-V-4.6 samples at 1 fps, with uniform
subsampling to at most 128 frames for longer clips. LLaVA-OneVision uses
the exact Hugging Face checkpoint's video processor; Dispider retains its
method-specific video encoder.

Rows marked $S$ use the selected-clip configuration: Qwen3.5-4B receives
up to 64 sampled frames, and Qwen3.8-27B uses the selected-clip
response configuration. The $S$ marker identifies this visual-input
configuration in the tables.
Table~\ref{tab:interaction-ovo-full} provides task breakdowns from the
same runs as the corresponding summary rows.

\begin{table}[htbp]
\centering\small
\caption{OVO content-task breakdown (\%). Counts refer to full-release probes. REC is separate from the perception and backward macros.}
\label{tab:interaction-ovo-full}
\setlength{\tabcolsep}{3pt}
\resizebox{\linewidth}{!}{%
\begin{tabular}{@{}lrrrrrrrrrr@{}}
\toprule
& \multicolumn{6}{c}{Perception} & \multicolumn{3}{c}{Backward tracing} & \\
\cmidrule(lr){2-7}\cmidrule(lr){8-10}
Model & OCR & ACR & ATR & STU & FPD & OJR & EPM & ASI & HLD & REC \\
Probes & 149 & 109 & 116 & 178 & 101 & 184 & 297 & 148 & 186 & 698 \\
\midrule
MiniCPM-V-4.6 & 77.18 & 55.05 & 65.52 & 46.63 & 71.29 & 58.15 & 56.23 & 52.70 & 26.88 & 28.22 \\
LLaVA-OneVision-7B & 69.13 & 57.80 & 72.41 & 53.37 & 71.29 & 61.96 & 54.88 & 55.41 & 22.04 & 25.93 \\
LiveStar-8B & 60.40 & 51.38 & 63.79 & 43.82 & 71.29 & 59.24 & 54.55 & 61.49 & 15.05 & 0.00 \\
Dispider & 57.05 & 49.54 & 62.07 & 44.38 & 61.39 & 51.63 & 48.48 & 55.41 & 4.30 & 18.19 \\
\midrule
Qwen3.5-4B$^{S}$ & 73.83 & 61.47 & 71.55 & 57.30 & 75.25 & 62.50 & 57.58 & 66.22 & 37.63 & 32.52 \\
Qwen3.8-27B$^{S}$ & 77.18 & 76.15 & 80.17 & 63.48 & 80.20 & 70.65 & 64.98 & 78.38 & 52.69 & 43.12 \\
\bottomrule
\end{tabular}%
}
\end{table}

\subsection{Omni-MMI: Prediction and Multi-Turn Consistency}
\label{app:interaction-omni}

AP evaluates action prediction with 200 single-answer records. SG evaluates
changing scene states across 300 records and 704 answer events; MD evaluates
multi-turn dependencies across 300 records and 786 answer events.
The model receives visual observations and the task's question context at
the supplied query points and generates text answers. The reported runs
use no native audio or ASR.

We use the released OpenAI judging prompt with GPT-4-turbo to assess
semantic correctness against the reference answer. The yes/no judgment,
rather than a threshold on the accompanying numeric rating, determines
correctness. For SG and MD, a record is correct only when every answer
is correct:
\begin{equation}
A_{\rm all}=\frac{100}{N}\sum_{i=1}^{N}\prod_{j=1}^{m_i}c_{ij},
\qquad c_{ij}\in\{0,1\}.
\end{equation}
Here $m_i$ is the number of answers in record $i$; AP is the single-answer
case. This distinguishes a correct isolated response from consistency
throughout an interaction. The accuracies in
Table~\ref{tab:interaction-content-main} use all 800 records and
1,690 answer events per evaluated model.

\subsection{StreamBench v0.3: Causal Question Answering}
\label{app:interaction-stream}

StreamBench v0.3~\citep{ppb_stream2025} is distinct from StreamingBench's Proactive Output task.
At each supplied question time, the model receives up to 16 visual samples
from the observed video prefix, with denser coverage of the most recent
20 seconds, ASR from up to 20 seconds of recent audio, and its own earlier
question--answer history. Selected frame timestamps and retained dialogue
turns cannot exceed the current query time; reference answers never enter
the history. Models generate text with greedy decoding and a 128-token
limit. This causal-prefix adaptation uses the full released question set,
rather than the full-video input of some released comparison scripts.

The six tasks are object search (OS), long-term memory search (LM),
short-term memory search (SM), conversational interaction (CI),
knowledge-based question answering (KG), and simple factual questions
(SF). The released Meta-Llama-3-8B-Instruct judge determines semantic
yes/no correctness. Task accuracy averages these judgments over all
questions in that task. We summarize search, memory, and interaction by
\begin{equation}
A_{\rm core}=\tfrac14(A_{\rm OS}+A_{\rm LM}+A_{\rm SM}+A_{\rm CI}).
\end{equation}
This selected four-task macro is not an official whole-benchmark score.
Table~\ref{tab:interaction-stream-full} retains all six task accuracies,
including KG and SF.

\begin{table}[htbp]
\centering\small
\caption{StreamBench v0.3 semantic accuracy under the causal visual-plus-ASR protocol (\%). All models cover the same 307 records and 1,838 questions. Core macro averages OS, LM, SM, and CI before rounding.}
\label{tab:interaction-stream-full}
\setlength{\tabcolsep}{4pt}
\begin{tabular}{@{}lrrrrrrr@{}}
\toprule
Model & OS & LM & SM & CI & KG & SF & Core macro \\
Questions & 299 & 303 & 339 & 297 & 298 & 302 & \\
\midrule
Qwen3-VL-8B & 35.79 & 43.56 & 54.28 & 65.99 & 72.82 & 73.51 & 49.91 \\
Qwen3.5-4B & 37.12 & 43.23 & 53.10 & 68.69 & 75.84 & 71.85 & 50.54 \\
MiniCPM-V-4.6 & 53.18 & 54.13 & 60.47 & 73.74 & 73.15 & 72.19 & 60.38 \\
LLaVA-OneVision-7B & 35.45 & 47.85 & 53.39 & 72.39 & 83.56 & 73.84 & 52.27 \\
LiveStar-8B & 32.11 & 36.30 & 48.97 & 62.63 & 44.30 & 67.55 & 45.00 \\
\bottomrule
\end{tabular}
\end{table}

\FloatBarrier

\section{Role-Conditioned Streaming Evaluation}
\label{app:role-proactive-protocol}

This evaluation asks whether a system can provide timely, relevant responses
when an explicit question is replaced by standing responsibilities.
The paired conditions share source videos and reference annotations,
so their comparison isolates the change in task conditioning.
Timing and response quality are measured from the same generated trajectories,
as reported in Tables~\ref{tab:timing-high-frequency}
and~\ref{tab:role-high-frequency-outcome}.

\subsection{Dataset Construction and Paired Conditions}

We construct a 200-video subset from the EGO and WEB portions of
ProactiveVideoQA, selecting 100 distinct parent videos from each domain.
Selection is based on source content and its compatibility with a shared
role catalog, without using the evaluated models' scores.
The subset is purposefully selected rather than a random sample of the
complete benchmark.

In the \emph{Explicit} condition, the system receives the original question.
In the \emph{Role} condition, that question is replaced by an identity and
general responsibilities appropriate to the scene.
The original videos, reference answers, and \texttt{reply\_timespan}
intervals are retained in both conditions.
No new intervention windows, recipient identities, or ordering constraints
are added.
Reference answers and timing annotations are used only for scoring and
are never included in model input.
Each source clip is replayed from beginning to end, rather than cropped
around its annotated answers.

Table~\ref{tab:role-conditioning-examples} shows two actual prompt pairs.
The cooking role covers several responsibilities without naming avocado
or prescribing its preparation steps; the driving role does not identify
the white car or its expected behavior.
This changes the information supplied to the system while keeping the
scoring targets fixed.

\begin{table}[htbp]
\centering\small
\caption{Examples of the Explicit-to-Role adaptation. Questions and role
prompts are reproduced from the constructed subset; videos, answers, and
reference time intervals are unchanged.}
\label{tab:role-conditioning-examples}
\begin{tabular}{@{}p{0.28\linewidth}p{0.67\linewidth}@{}}
\toprule
Original question (Explicit) & Standing responsibilities (Role) \\
\midrule
What are the different steps involved in preparing avocado in this process?
&
You are a cooking assistant supporting everyday food preparation.
Attend to progress, ingredient and utensil use, and safety.
Offer brief guidance, useful updates, or reminders when helpful,
without repetitive interruptions. \\
\addlinespace
What is happening with the white car?
&
You are a driving assistant helping a driver understand surrounding traffic.
Attend to road conditions, vehicle movements, and safe passage.
Provide concise, timely information without continuously narrating normal driving. \\
\bottomrule
\end{tabular}
\end{table}

The retained annotations provide a common reference for comparing the
two conditions; they are not an exhaustive annotation of every useful
role-consistent intervention.
Accordingly, an unmatched response is an error relative to this reference,
not necessarily evidence that it would be unhelpful in an unconstrained setting.

\subsection{Causal Inputs and System Interfaces}

For the rolling-window VLM interface, observations are updated every
200\,ms.
Each request uses only frames available at its start: the most recent
five seconds are sampled at 5\,Hz and the preceding 25 seconds at 1\,Hz,
with duplicate boundary frames removed.
Frames are resized to a maximum side length of 512 pixels.
This retains approximately 50 frames in a 30-second window rather than
sending all 150 observations at every update.

Speech is supplied as causal ASR text using Qwen3-ASR-1.7B.
The shared ASR schedule records when each finalized utterance becomes
available, including endpoint detection and recognition delay.
An utterance enters a request only after that availability time and only
if its source interval lies within the retained context.
The dialogue history contains at most the five latest completed system
responses and their timestamps, with a 1,024-token limit; older entries
are removed first.
Reference replies, future transcripts, and unfinished generated responses
are not inserted into this history.

A standalone VLM generates a natural-language response or the exact
control marker \texttt{NO\_UPDATE}.
The latter denotes silence; other nonempty visible text is treated as
a response rather than retrospectively reinterpreted as silence.
Empty output, incomplete control markers, and transport failures are
not valid silence.
Potential control prefixes are buffered until their interpretation is
determined, and visible-response time is measured when text is actually released.

For the matched dual-model comparison, the unmodified Qwen3.5-4B and
StreamingJev control the same Qwen3.8-27B Worker.
The pair uses the same input and scheduling budgets.
The Timing Model emits \texttt{response}/\texttt{silence}; the Worker
generates response content when triggered.
Timing and response generation operate asynchronously.
A trigger freezes the currently available Worker context.
While the Worker is busy, further Timing decisions are logged but do
not create a second queued response; another response requires a new
trigger after completion.

LiveStar~\citep{yang2025livestar} and Dispider~\citep{qian2025dispider} retain their native observation schedules, memory,
and trigger/response mechanisms rather than being reduced to the
rolling-window VLM interface.
Their adapters use visual input without adding ASR to the native
decision paths.
Thus, 200\,ms specifies our observation-update setting, not a requirement
that every compared method ingest frames or finish decisions at 5\,Hz.

\subsection{Timing and Response-Quality Measures}

Let $I_j=[a_j,b_j]$ denote an original reference reply interval and
$t_i$ the first visible-text time of a generated response.
Timing evaluation constructs a deterministic maximum-cardinality
one-to-one matching between responses and intervals satisfying
$t_i\in I_j$.
Matched pairs are true positives, unmatched responses are false positives,
and unmatched reference intervals are false negatives.
We report
\[
F_1=\frac{2TP}{2TP+FP+FN},
\qquad
\mathrm{FP/min}=\frac{FP}{T/60},
\]
where $T$ is the evaluated video duration in seconds.
The main comparison averages these measures over episodes.
These timing measures do not inspect answer semantics.
LiveStar's Omni-MMI PA/PT scores are omitted because its native response
interface is incompatible with the required decision format.
Omni-MMI PA/PT is a separate task-conditioned evaluation, not a
role-transformed part of this subset.

\paragraph{Outcome.}
Each original reference answer defines one response target.
For target $j$, we gather generated responses whose first visible text
falls within $I_j$ and ask whether they jointly convey all essential
reference information, including the specified people or objects.
Paraphrases are accepted and repetition earns no additional credit.
Complete coverage scores one; partial, incorrect, or absent coverage
scores zero.
No identity or ordering requirement absent from the reference is introduced.
For an episode with $m$ reference targets,
\[
S_{\mathrm{Outcome}}=\frac{1}{m}\sum_{j=1}^{m}c_j,
\qquad c_j\in\{0,1\}.
\]
Scores are first averaged within each episode and then across episodes.
This measures delivered verbal content, not completion of a physical action.

\paragraph{PAUC.}
We follow the ProactiveVideoQA metric and official scoring
procedure~\citep{wang2025proactivevideoqa}, including cumulative response
deduplication, semantic judgments, time integration, and aggregation
across reference reply turns. For a reference turn $g$ with assistance
interval $[a_g,b_g]$, let $t_1<\cdots<t_m$ be the response timestamps
within that interval. At each $t_i$, GPT-4.1 evaluates the accumulated
response content against the reference answer, assigning
$s_i\in\{0,1,2\}$ for completely incorrect, partially correct, and mostly
correct content, respectively. To balance timeliness and correctness,
the metric transforms each timestamp to
\begin{equation}
u_i=a_g+(1-\omega)(t_i-a_g),\qquad \omega=0.5.
\end{equation}
Starting from a score of $0.5$, it integrates the resulting
piecewise-constant score curve and normalizes by the maximum possible
area $2(b_g-a_g)$:
\begin{equation}
\mathrm{PAUC}_g=
\frac{0.5(u_1-a_g)+\sum_{i=1}^{m-1}s_i(u_{i+1}-u_i)
+s_m(b_g-u_m)}{2(b_g-a_g)}.
\end{equation}
If there is no response within the interval, the official initial score
gives $\mathrm{PAUC}_g=0.5/2=0.25$. The reported score is the mean over
reference reply turns, following the official implementation. Thus,
PAUC rewards correct information delivered earlier within the assistance
window and is distinct from binary coverage accuracy.
Outcome retains its own binary coverage criterion. Both semantic
evaluations use GPT-4.1 with model identities hidden from the judge:
the official prompt for PAUC and a fixed coverage prompt for Outcome.

All these measures use the same self-triggered trajectories.
There is no second generation pass at reference trigger times.
A reply beginning within a reference interval may be evaluated for content
even if generation finishes later; its completion time is recorded separately.

\subsection{Runtime Configuration and Measurement}
\label{app:condition-runtime}

For the rolling-window interface, inference is single-stream and does
not accumulate a queue of obsolete requests.
Observations continue to enter the buffer while generation is in progress.
When the system becomes available, it processes the latest eligible
context instead of replaying missed inference opportunities.
Video end stops new requests; in-flight output keeps its actual timestamps
and is not backdated.

\begin{table}[htbp]
\centering\small
\caption{Runtime measures accompanying the paired evaluation.
Response latency and completed decision frequency describe different
parts of the streaming workload.}
\label{tab:role-runtime-definitions}
\begin{tabular}{@{}p{0.23\linewidth}p{0.72\linewidth}@{}}
\toprule
Measure & Definition \\
\midrule
Response latency &
Time from the start of the processing cycle that produces a response
to its first visible text, including input preparation and scheduling.
For the dual-model system, the cycle starts with the triggering Timing request
and ends with the Worker's first visible text. \\
\addlinespace
Decision frequency &
Number of completed intervention decisions, including silence decisions,
divided by effective playback duration.
For the dual-model system, decisions are counted at the Timing Model.
Completions after video end are recorded separately. \\
\addlinespace
Completion latency &
Time from the same processing-cycle start to the last response text;
distinct from first-text latency and reference-time error. \\
\bottomrule
\end{tabular}
\end{table}

The main response table reports mean first-text latency over emitted
responses. Table~\ref{tab:role-decision-frequency} reports decision
frequency as total completed decisions divided by total effective playback
time. ProAgent's 200\,ms observation interval corresponds to a 5\,Hz input
cadence; its measured completed decision frequencies are 4.723\,Hz under
Explicit and 4.754\,Hz under Role.
An episode with no visible response does not receive zero response latency.
Input frame rate, token-generation rate, and inverse response latency
are not substitutes for decision frequency.

\begin{table}[!ht]
\centering\small
\caption{\textbf{Completed decision frequency in the paired evaluation.}
Decision Hz counts completed intervention decisions per second of effective
playback, including silence decisions. For the dual-model systems,
decisions are counted at the Timing Model.}
\label{tab:role-decision-frequency}
\setlength{\tabcolsep}{7pt}
\begin{tabular}{@{}lrr@{}}
\toprule
System & Explicit (Hz) & Role (Hz) \\
\midrule
\multicolumn{3}{@{}l}{\textit{General-purpose: open weights}}\\
Qwen3.5-4B & 0.4687 & 0.4780 \\
Qwen3-VL-8B-Instruct & 0.6215 & 0.6420 \\
MiniCPM-V-4.6 & 0.8196 & 0.8297 \\
\midrule
\multicolumn{3}{@{}l}{\textit{General-purpose: closed source}}\\
GPT-6 & 0.0692 & 0.0610 \\
Gemini & 0.0671 & 0.0661 \\
Claude & 0.0710 & 0.0774 \\
\midrule
\multicolumn{3}{@{}l}{\textit{Streaming/proactive methods}}\\
LiveStar-8B & 0.3411 & 0.3238 \\
Dispider & 0.7129 & 0.7177 \\
\midrule
\multicolumn{3}{@{}l}{\textit{Matched Timing Models + shared Worker}}\\
Qwen3.5-4B Base + Worker & \textbf{4.812} & \textbf{4.792} \\
StreamingJev + Worker (Ours) & 4.723 & 4.754 \\
\bottomrule
\end{tabular}
\end{table}

Open-weight standalone VLMs use native BF16 inference on an A800 80\,GB
GPU; the matched dual-model deployment uses a separate GPU for each
model, excluding shared ASR preparation.
Model loading, compilation, and warm-up are recorded separately from
replay.
Decoding settings, output limits, and enabled inference optimizations
are retained with each measured configuration rather than assumed to
be identical across backends.
Both matched Timing/Worker configurations disable thinking;
other models disable it where the serving interface supports this setting.
A response reaching its token limit is retained for scoring and marked
as truncated.

Remote-service measurements include network and service time and are
reported separately from local-GPU systems.
These are aggregate response times: the comparison does not separately
attribute latency to internal reasoning, network transfer, or service processing.
Where the declared fault-pause policy applies, service-recovery intervals
are excluded from effective playback time, while raw wall-clock time
and excluded duration are retained.
Ordinary slow inference remains part of playback and latency.

\FloatBarrier

\Needspace{10\baselineskip}
\section{Embodied Evaluation Protocol and Case Studies}
\label{app:physical-protocols}

\subsection{Task Categories and Information Access}

The evaluation consists of 60 cases, divided equally among Guidance, Warning, and Interaction (Table~\ref{tab:physical-task-allocation}). Each case specifies an evolving physical situation, the agent's standing responsibilities, and the communication or device functions available to it. The intervention reference comprises a reference time and required events with associated people or objects and ordering constraints. These references are hidden from the system during inference.

\begin{table}[htbp]
\centering\small
\caption{Functional coverage of the embodied evaluation.}
\label{tab:physical-task-allocation}
\begin{tabular}{@{}lcp{0.60\linewidth}@{}}
\toprule
Category & Cases & Intervention function \\
\midrule
Guidance & 20 & Timely information, instruction, and assistance during an ongoing activity. \\
Warning & 20 & Correction or warning in response to emerging errors and risks. \\
Interaction & 20 & Recognition of human needs and sustained proactive engagement. \\
\midrule
Total & 60 & \\
\bottomrule
\end{tabular}
\end{table}

Standing responsibilities define the scope of assistance without disclosing the case-specific trigger or expected response. Event matching uses observable interaction or execution evidence: describing an intended physical action or receiving a function-call acknowledgment does not establish that the action occurred.

\subsection{Metric Definitions}

Let $t_{\rm start}$ denote the reference time at which an intervention is warranted and $t_{\rm trigger}$ the observed response initiation time. Time measures their absolute deviation:
\begin{equation}
L=|t_{\rm trigger}-t_{\rm start}|.
\end{equation}
Both early and late interventions increase this error.

Let $N_{\rm total}$ be the number of required evaluation events, including unexecuted events, and $N_{\rm match}$ the number matched in content and associated person or object. Outcome is
\begin{equation}
S_{\rm outcome}=\begin{cases}
N_{\rm match}/N_{\rm total}, & \text{if the prescribed execution order is respected},\\
0, & \text{otherwise}.
\end{cases}
\end{equation}
Only prescribed ordering constraints are evaluated; independent events are not assigned an artificial total order. Outcome is reported on a 0--1 scale, with higher scores indicating more complete assistance.

\subsection{Aggregation}

Results are summarized separately for each functional category. The main table reports mean Time and mean Outcome. Missed interventions are distinguished from zero-error measurements when computing Time.

\subsection{Qualitative Embodied Case Studies}
\label{app:embodied-case-studies}

\begin{figure}[!htbp]
    \centering
    \includegraphics[width=\linewidth]{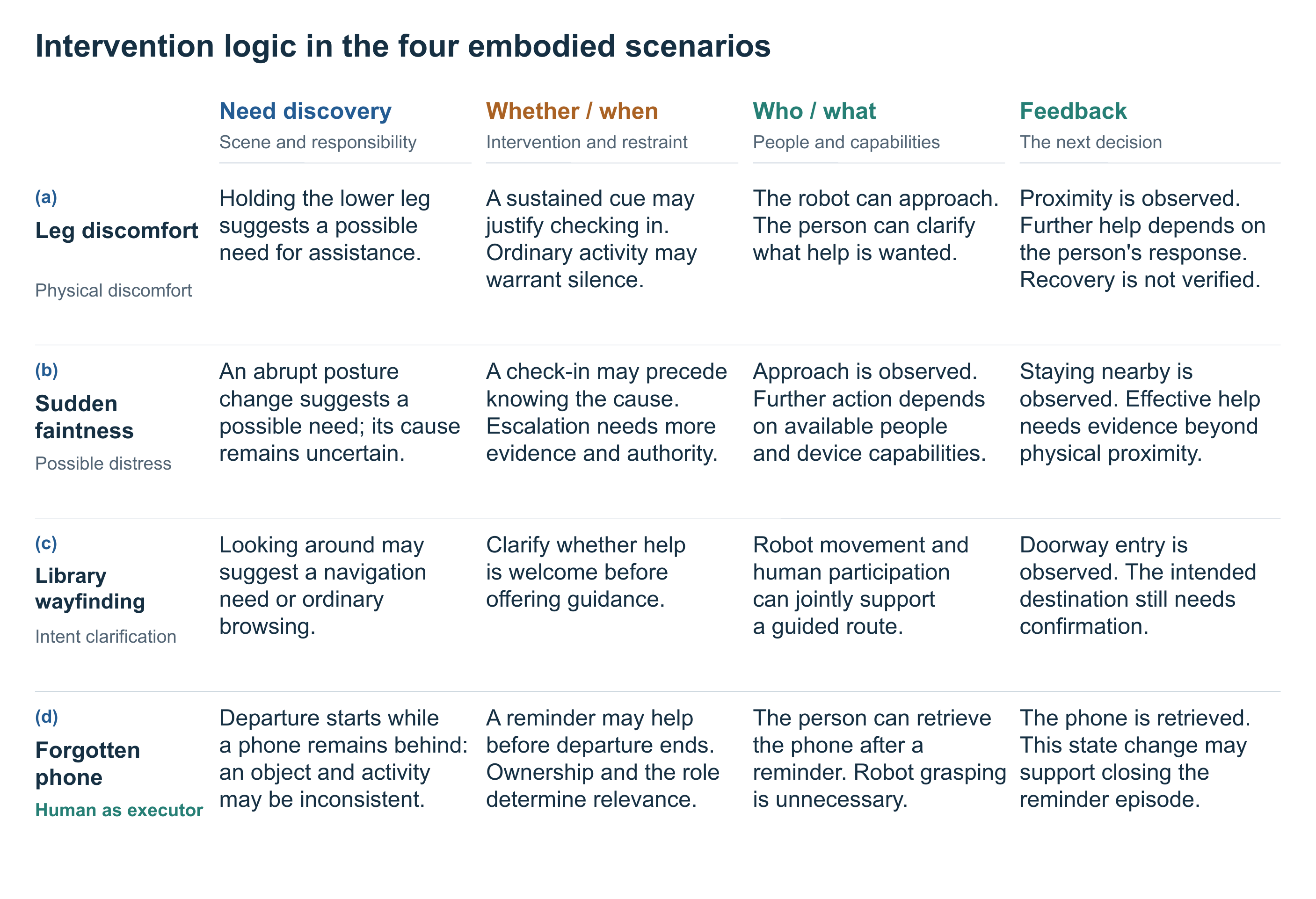}
    \caption{\textbf{Interpreting the cases through the ProPAI formulation.} The columns relate each scenario to a possible assistance need, whether and when intervention may be appropriate, who or what could execute a response, and how subsequent observations should inform the next decision. The phone case highlights the person as a physical-task executor: a reminder can support object retrieval without requiring robot manipulation. The final column distinguishes an observed state change from evidence needed to establish completion. The intervention logic provides a conceptual interpretation of the observed actions and states in Figure~\ref{fig:embodied-case-trajectories}.}
    \label{fig:embodied-case-interpretation}
\end{figure}

We examine four recorded scenarios to make the task-level decisions of ProPAI concrete. Figure~\ref{fig:embodied-case-trajectories} presents the visible interaction sequences, while Figure~\ref{fig:embodied-case-interpretation} explains how the situations relate to need discovery, intervention timing, available executors, and feedback. These recordings illustrate the system's interaction mechanisms; category-level scores are computed from the separate 60-case evaluation.

\paragraph{People as physical-task executors.}
The phone sequence illustrates why the response space includes both human-facing communication and device-facing actions. The person changes the physical state by retrieving the phone. An assistant could therefore help through a reminder, while the person supplies the manipulation capability. The visible retrieval provides an observable outcome for assessing whether the object has been recovered.

\paragraph{Feedback determines what remains unresolved.}
The cases illustrate the role of feedback in assessing task completion. Proximity can be observed when the robot approaches a person; subsequent observations and human feedback are needed to assess whether assistance has resolved the underlying need. Similarly, navigation progress and arrival at the intended destination represent distinct states. These distinctions inform whether an intervention should continue, change, or conclude.

\section{Efficiency Evaluation and Component Analysis}
\label{app:efficiency-protocols}

The efficiency comparison in Table~\ref{tab:efficiency-main-draft}
examines how Incremental Caching, Dynamic Token Dropping, and Prefix-constrained Decoding reduce streaming decision latency
and peak GPU memory usage.
This appendix describes the cumulative configurations, measurement
definitions, interpretation of the reported results, implementation details, and effects of the DTD budgets.

\subsection{Workload and Cumulative Configurations}
\label{app:efficiency-workload}

We evaluate the trained StreamingJev Timing Model on 398 temporal
sequences drawn from EgoLife, Kinetics, CharadesEgo, XD-Violence,
TimeLens, EgoExoLearn, and UR-Fall. The collection comprises 332 unique
source video files and 3,370 seconds of video, yielding 3,768 timing
decisions per configuration. Each update introduces one new frame sampled
at 1\,Hz, and retains the complete append-only history of its sequence.
Sequences contain 1--17 frames. Inputs use dynamic image resolution with
a processor pixel range of 65,536--262,144; audio and ASR text are excluded.

All configurations use the same trained model on a single NVIDIA
A800-SXM4-80GB GPU with BF16 precision, SDPA attention, and batch size 1.
Only one stream and one configuration run at a time. The inference
backend is a custom Transformers adapter using Transformers 5.14.1,
PyTorch 2.10.0, and FLA operators supplied by vLLM 0.18.1.

We compare four cumulative configurations.
\emph{Full recomputation} processes the retained multimodal context
at each update. 
\emph{Incremental Caching} processes only newly appended frames and
reuses historical post-vision-encoder features and compatible
language-model KV states.
\emph{DTD} removes redundant current-frame visual tokens by comparing
features at the same token-grid indices in the preceding frame when
their visual grids match.
\emph{Prefix-constrained Decoding} further restricts generation to legal decision
prefixes and stops when the binary decision is uniquely determined.
The mechanisms are described in Section~\ref{sec:timing-efficiency}.

Each row measures the additional benefit of its component in this
ordering, with DTD fixed at $\tau=0.5$. All configurations receive
identical images, text, and frozen causal histories. Historical assistant
outputs are taken from an initial full-recomputation pass, so the
comparison holds interaction history fixed rather than allowing each
configuration's outputs to alter its subsequent inputs.

Before measurement, each configuration is warmed up on the first,
a middle, and a longest sequence; warm-up measurements are excluded.
The results come from one complete paired run with configurations
interleaved in groups. Each sequence starts with a fresh session.
Because histories contain at most 17 frames, this experiment does not
exercise history eviction or sliding-window shifts. Cache validity
and reset rules are described in Section~\ref{app:engine-efficiency-design}.

\subsection{Latency, Memory, and Decision Agreement}

Let $a_i$ be the synchronized inference start time after image decoding
and preprocessing for update $i$, and $b_i$ the synchronized completion
time of its timing decision. Inference latency is
\begin{equation}
\ell_i=b_i-a_i.
\end{equation}
We synchronize the GPU before starting a high-resolution wall-clock timer
and again after inference completes. Latency excludes image decoding and
preprocessing. P50 and P95 are computed over all 3,768 update latencies
for each configuration, rather than averaging per-video quantiles.

Memory is peak PyTorch allocated GPU memory, including model weights,
activations, and caches. Peak statistics are reset for each sequence,
and the reported value is the maximum across sequences, in GB.

Decision agreement compares the configuration's output $\hat y_i$ with
the full-recomputation output $y_i^{\mathrm{full}}$ at the same update
in the paired run:
\begin{equation}
\mathrm{Agree}=\frac{100}{N}\sum_{i=1}^{N}
\mathbf{1}[\hat y_i=y_i^{\mathrm{full}}],\qquad N=3{,}768.
\end{equation}
The DTD and complete configurations each match 3,717 reference decisions,
giving 98.65\% agreement. Agreement measures preservation of the baseline
decision, not task accuracy. All evaluated updates produced valid decisions.

\subsection{Component Effects}

As shown in Table~\ref{tab:efficiency-main-draft}, Incremental Caching
reduces P50 latency from 346.84 to 240.26\,ms while retaining 100.00\%
decision agreement. Adding DTD further reduces P50 to 214.16\,ms,
with agreement of 98.65\%. Prefix-constrained Decoding lowers P50 to
147.05\,ms without a further change in the reported agreement.
Across the cumulative configurations, P95 decreases from 512.63 to
251.77\,ms and peak memory from 9.8544 to 9.50\,GB.

The complete configuration reduces median latency by approximately
57.6\% and P95 by 50.9\% relative to full recomputation.
These results describe the cumulative execution path rather than
order-independent effects of the individual mechanisms.
These measurements isolate inference cost on the fixed 1\,Hz workload.
The 200\,ms observation cadence and completed decision frequencies in
the paired interaction evaluation are reported separately in
Appendix~\ref{app:condition-runtime}.

\subsection{Implementation Details.}
\paragraph{Incremental Caching.}
We implement the three mechanisms as independently switchable modules
around the trained StreamingJev Timing Model, keeping its weights
fixed across configurations. Each stream
is assigned an isolated session state containing the ordered image
identifiers, full post-vision-encoder features of previously observed
frames, and the language model's KV cache for the stable prompt prefix.
At each update, only newly appended frames undergo image decoding,
resizing, normalization, and visual encoding. The prompt tokens and
multimodal position indices are reconstructed to verify prefix
compatibility, after which the language model evaluates only the
uncached suffix. The mutable generation suffix is excluded from the
persistent prefix state. Image reuse requires an append-only frame history. Removing, replacing,
or reordering a historical frame causes the session to be rebased.
Likewise, a mismatch in cached token IDs, position IDs, or KV-cache
length invalidates the language-model prefix cache. Changes to the model,
processor, chat template, or DTD configuration require a new session.
Thus, no cached state is shared across unrelated streams.

\paragraph{Dynamic Token Dropping.}
Let $\mathbf{z}_{t,j}$ and $\mathbf{z}_{t-1,j}$ denote the post-vision-
encoder features at token-grid index $j$ in the current and immediately
preceding frames. When the two frames have matching visual grids and
feature shapes, DTD computes their cosine similarity in FP32:
\begin{equation}
s_{t,j} =
\frac{\mathbf{z}_{t,j}^{\top}\mathbf{z}_{t-1,j}}
{\lVert\mathbf{z}_{t,j}\rVert_2
 \lVert\mathbf{z}_{t-1,j}\rVert_2}.
\end{equation}
A current-frame token is retained when $s_{t,j}\leq\tau$ and removed
when $s_{t,j}>\tau$, where $\tau$ is the DTD threshold. At least
$K_{\min}=1$ token with the lowest similarity is retained. The first
frame and frames with incompatible grids are kept in full, and
$\tau=1$ defines the identity configuration. The keep mask is applied consistently to the visual placeholders,
embeddings, and position indices before language-model prefill.
Retained tokens preserve their original multimodal rotary positions.
The full, unpruned current-frame features are cached as the reference
for the next frame, preventing pruning errors from accumulating across
time. DTD affects only newly appended tokens and never modifies
historical tokens already stored in the KV cache.

\paragraph{Prefix-constrained Decoding.}
The two legal outputs, \texttt{<silence>} and \texttt{<response>}, are
tokenized and organized as a prefix trie, with a maximum generation
length of eight tokens. At each decoding step, the
output projection is evaluated only for tokens that extend a legal
prefix. Once the remaining prefix is compatible with exactly one
decision, the corresponding canonical token sequence is emitted without
additional decoder forward passes. This reduces both full-vocabulary
projection and autoregressive model execution while preventing invalid
outputs.

\subsection{Effects of the DTD Threshold.}\label{app:dtd-threshold}
DTD removes a current-frame visual token when its cosine similarity
to the token at the same grid location in the preceding frame exceeds
the threshold $\tau$. A lower $\tau$ therefore produces more aggressive
pruning, whereas $\tau=1.0$ is an identity configuration that retains
all visual tokens. We sweep $\tau$ from 0.5 to 1.0 while holding the
checkpoint, input histories, and inference backend fixed. Apart from
DTD, incremental caching is the only optimization enabled throughout
the sweep; prefix-constrained decoding and all other optimization
modules are disabled. Lowering $\tau$ from 1.0 to 0.5 reduces the
retained-token ratio from 100.00\% to 16.37\%, removing 83.63\% of
current-frame visual tokens at the most aggressive setting.

\begin{table}[t]
\centering
\small
\caption{\textbf{Sensitivity to the DTD threshold.}
Retention denotes the percentage of current-frame visual tokens retained.
Agree. is decision agreement with full recomputation. Mem. is peak live
GPU allocation in GB.}
\label{tab:dtd-threshold-sweep}
\setlength{\tabcolsep}{3.5pt}
\begin{adjustbox}{max width=\linewidth}
\begin{tabular}{@{}crrrrr@{}}
\toprule
$\tau$
& Retention (\%) $\downarrow$
& P50 (ms) $\downarrow$
& P95 (ms) $\downarrow$
& Mem. (GB) $\downarrow$
& Agree. (\%) $\uparrow$ \\
\midrule
1.0 & 100.00 & 240.26 & 270.06 & 9.63 & 100.00 \\
0.9 &  67.76 & 214.16 & 269.73 & 9.50 &  98.65 \\
0.8 &  51.67 & 211.25 & 252.12 & 9.44 &  97.08 \\
0.7 &  37.82 & 207.50 & 249.89 & 9.38 &  96.13 \\
0.6 &  25.34 & 202.84 & 253.02 & 9.33 &  94.29 \\
0.5 &  16.37 & 197.49 & 247.16 & 9.29 &  93.42 \\
\bottomrule
\end{tabular}
\end{adjustbox}
\end{table}

DTD provides a clear reduction in median latency, while its effects on
tail latency and memory are more modest. Relative to $\tau=1.0$,
$\tau=0.5$ reduces P50 latency by 17.8\%, from 240.26 to 197.49\,ms,
and P95 latency by 8.5\%, from 270.06 to 247.16\,ms. Peak allocated
memory decreases by 3.5\%, from 9.63 to 9.29\,GB. This difference is
expected because DTD is applied after the vision encoder: it reduces
language-model prefill and subsequent KV-cache growth, but does not
avoid visual-encoder computation. Moreover, model weights account for
most of the measured GPU memory.

More aggressive pruning introduces a corresponding tradeoff in decision
consistency. Agreement with full recomputation decreases from 100.00\%
at $\tau=1.0$ to 93.42\% at $\tau=0.5$. The more conservative
$\tau=0.9$ setting retains 67.76\% of visual tokens, reduces P50 latency
by 10.9\%, and preserves 98.65\% decision agreement. The deployment
threshold can therefore be selected according to the required balance
between latency and decision consistency. Because each threshold was
evaluated in a single full pass, small non-monotonic differences,
particularly in P95 latency, should be treated as descriptive rather
than statistically significant.

\begingroup

\section{System Implementation and Training Details}
\label{app:engine-extended-design}

ProAgent connects continuous observation to intervention through three
parts of its implementation: the runtime that maintains context and
coordinates the two models, the data pipeline that supplies timing
supervision, and the capability interface that turns responses into
communication or device actions. This appendix follows that path,
complementing the architecture in Section~\ref{sec:system}.

\subsection{Runtime, Context, and Memory}
\label{app:engine-runtime}
\label{app:engine-memory-design}
\label{app:engine-efficiency-design}

\paragraph{Continuous observation and asynchronous response.}
The Timing and Response Models share an observation source but maintain
separate model-dependent states. The runtime makes recent observations,
standing responsibilities, available capabilities, interaction history,
and ongoing intervention state available to each model.
The Timing Model decides whether output should be active; the Response
Model determines the response content and executor. Response-context
preparation and timing prediction can proceed asynchronously, while
generation remains trigger-gated. The evaluated replay and busy-state
handling are specified in Appendix~\ref{app:role-proactive-protocol}.

\paragraph{Context availability and state reuse.}
A context item becomes usable when it is available to the system, not
merely when the underlying event occurred. In particular, a transcript
finalized after a decision cannot be inserted into that earlier decision.
Observation timestamps, derived-information availability, and completed
response times therefore have distinct roles in constructing causal inputs.
The retained context and sampling settings of each experiment are given
in its evaluation protocol rather than assumed to be universal.

Each model's cache corresponds to its own weights, prompt, retained
inputs, and position assignments. Sharing observations does not imply
sharing KV states across the two models. Reuse requires an unchanged
prefix; removing old context, changing its representation, or resetting
a session requires updating or invalidating the affected cached states.
Appendix~\ref{app:efficiency-protocols} specifies the efficiency comparison
and the component-level measurements.

\paragraph{Interaction history and unfinished tasks.}
Recent perceptual context explains what has just changed, whereas
interaction history records what the system has already said or attempted.
Ongoing intervention state serves a different purpose: it distinguishes
a need that remains unresolved from a response currently being generated
or a device operation still in progress. These are runtime distinctions,
not additional classes in the Timing Model's binary output.
They prevent each observation update from being treated as an unrelated
opportunity to start the same intervention.

Longer-lived task, episodic, and user/environment memory can extend this
context beyond the current observation window. Such memory would retain
the source and validity of a belief, distinguish uncertain inferences
from explicit instructions, and revise stale information when new evidence
arrives. Learning when to retrieve, correct, or forget these records is
a further design question, developed in
Section~\ref{app:engine-challenge-memory}.

\paragraph{Output ownership and execution feedback.}
Releasing the conversational output channel does not cancel a physical
operation already dispatched. Likewise, an accepted function call
establishes that execution was requested, not that the task succeeded.
Returned status and subsequent observations inform whether assistance
should continue, change, or end; cancellation and recovery use operations
supported by the executor. Coordination of concurrent tasks can build
on these distinctions by tracking resource commitments and linking retries
to the original request, rather than issuing independent duplicate actions.

\subsection{Timing Supervision and Training}
\label{app:engine-learning-design}
\label{app:rev-data}

\paragraph{Training data.}
The training corpus contains 224,268 multi-turn decision sequences from
20,258 source videos, with 2,058,267 supervised decision turns.
Each sequence pairs a task or role context with 1--17 successive visual
observations and their decision labels. Table~\ref{tab:rev-data} reports
the source distribution after construction, filtering, and deduplication.
Explicit-condition supervision accounts for 98,730 sequences (44.02\%);
role- and anomaly-conditioned supervision accounts for 125,538 (55.98\%),
comprising 91,864 role sequences, 31,394 confirmed-silence anomaly
sequences, and 2,280 anomaly-onset sequences.

\begin{table}[!htbp]
\centering\small
\caption{\textbf{Source distribution of the Timing Model training corpus.}
Counts refer to multi-turn decision sequences; percentages use the full
training corpus as the denominator.}
\label{tab:rev-data}
\setlength{\tabcolsep}{8pt}
\begin{tabular}{@{}lrr@{}}
\toprule
Source & Training sequences & Share (\%) \\
\midrule
COIN & 58,811 & 26.22 \\
HoloAssist & 41,920 & 18.69 \\
Ego-Exo4D & 38,037 & 16.96 \\
Anomaly videos & 33,674 & 15.02 \\
HACS & 22,322 & 9.95 \\
YouCook2 & 17,673 & 7.88 \\
EgoIT-99K & 6,253 & 2.79 \\
QEVD-FIT-COACH & 5,578 & 2.49 \\
\midrule
Total & 224,268 & 100.00 \\
\bottomrule
\end{tabular}
\end{table}

We preserve the source datasets' training and validation partitions
when constructing sequences. The validation split contains 46,573
sequences from 2,958 videos, with 421,287 supervised decision turns.
The two splits are disjoint at the source-video and event levels.
The training corpus is disjoint from all evaluation benchmarks used
in this paper, including OVO-Bench, Omni-MMI, StreamingBench, and
StreamBench.

\paragraph{Causal sequence construction.}
Offline annotation uses complete recordings to locate event context
and response windows; each training input contains only observations
available by its decision time. Regular sequences sample one frame per
second for up to 17 observations, covering approximately 16 seconds.
Subsecond events without a valid integer-second positive observation
retain a single observation at their native timestamp. The native
Qwen3.5 processor encodes images with a maximum area of 230,400 pixels.

The first turn supplies the task or role prompt, the current image, and
the output state; subsequent turns append new observations and states.
Teacher forcing retains preceding ground-truth decisions, and each
output state is updated causally from the preceding decision, starting
from an idle state. Training uses images and task/state text, with
supervision on decision labels. Causal ASR used in evaluation is
specified separately in Appendix~\ref{app:role-proactive-protocol}.
Unknown intervals are excluded, and silence sequences do not bridge
unknown gaps.

\paragraph{Optimization.}
We train StreamingJev from Qwen3.5-4B for three epochs, updating all
language-model parameters and the native output head while freezing
the visual encoder and visual merger. Table~\ref{tab:engine-training-settings}
summarizes the configuration. The model retains its native vocabulary:
\texttt{<response>} and \texttt{<silence>} are encoded as three and four
tokens, respectively. Their meanings depend on the output state:
START/WAIT when idle and CONTINUE/STOP when output is active.

\begin{table}[!htbp]
\centering\small
\caption{\textbf{Training configuration of StreamingJev.}}
\label{tab:engine-training-settings}
\begin{tabular}{@{}p{0.40\linewidth}p{0.55\linewidth}@{}}
\toprule
Setting & Value \\
\midrule
Backbone & Qwen3.5-4B \\
Trainable modules & Language model and native output head \\
Frozen modules & Visual encoder and visual merger \\
Epochs & 3 \\
Optimizer & AdamW \\
Peak learning rate & $2.5\times10^{-6}$ \\
Schedule / warm-up & Cosine / 5\% of updates \\
Adam parameters & $\beta_1=0.9$, $\beta_2=0.95$, $\epsilon=10^{-8}$ \\
Weight decay / gradient clipping & 0.01 / maximum norm 1.0 \\
Global batch size & 16 sequences (16 GPUs, one sequence per GPU) \\
Gradient accumulation & 1 \\
Maximum sequence length & 8,192 tokens \\
Maximum image area & 230,400 pixels \\
Precision / distributed strategy & BF16 / DeepSpeed ZeRO-2 \\
Random seed & 20260811 \\
\bottomrule
\end{tabular}
\end{table}

Equation~\ref{eq:engine-timing-loss} supervises all assistant decision
turns. Decision-text tokens receive weights of 0.8, 1.2, 1.0, and 1.0
for WAIT, START, CONTINUE, and STOP, respectively; turn-ending tokens
and retained newlines receive weight 1.0. Input tokens, padding, and
empty thinking prefixes are masked. The loss is normalized by the
number of valid supervised tokens in the global batch.
The corpus contains 974,144 response decisions (47.33\%) and
1,084,123 silence decisions (52.67\%).

\subsection{Response Generation and Capability Execution}
\label{app:engine-skills-design}
\label{app:physical-execution}
\label{app:pci-specification}
\label{app:response-model}

\paragraph{From an intervention decision to a response.}
The Response Model uses the observed situation, relevant people, available
devices, and previous interaction or execution results to decide whom to
address and what to do. An existing multimodal foundation model is adapted
through contextual prompting, without separately training a response
backbone. Human-facing text is passed to external TTS; device-facing
responses select callable operations and their arguments. A response may
combine communication with device actions.

\paragraph{Capability representation and discovery.}
Let $\mathcal P_t$ denote the available providers and $\mathcal K_p$ the
capabilities declared by provider $p$. The callable action space is
\begin{equation}
    \mathcal A^{\mathrm{call}}_t
    = \bigcup_{p\in\mathcal P_t}
      \{(i_p,n_k,v_k)\mid k\in\mathcal K_p\},
    \label{eq:engine-pci-action-space}
\end{equation}
where $i_p$ identifies a provider and $(n_k,v_k)$ the capability name
and version. Provider identity and capability identity remain separate:
a provider may offer several operations, and different providers may
offer the same semantic capability.

The local registry aggregates available capability descriptions and
presents them as function definitions to the Response Model.
A description specifies what an operation does and the schemas of its
arguments and return value. It does not require the model to know the
provider's private SDK, transport, or low-level controller.
Table~\ref{tab:engine-capability-fields} groups the representation by
its role in the interaction.

\begin{table}[htbp]
\centering\small
\caption{Fields connecting capability discovery, invocation, and execution feedback.}
\label{tab:engine-capability-fields}
\begin{tabular}{@{}p{0.19\linewidth}p{0.32\linewidth}p{0.43\linewidth}@{}}
\toprule
Representation & Fields & Role \\
\midrule
Provider descriptor &
\texttt{schema\_version}, \texttt{provider\_id}, \texttt{capabilities} &
Identify the representation and device provider, and enumerate its operations.
Optional \texttt{metadata} describes the device. \\
\addlinespace
Capability descriptor &
\texttt{name}, \texttt{version}, \texttt{description}, \texttt{input\_schema}, \texttt{output\_schema} &
Specify a namespaced operation, its meaning, accepted arguments, and returned values. \\
\addlinespace
Invocation &
\texttt{schema\_version}, \texttt{request\_id}, \texttt{provider\_id}, \texttt{capability}, \texttt{arguments} &
Select a provider and versioned capability; bind the operation's parameters. \\
\addlinespace
Return &
\texttt{schema\_version}, \texttt{request\_id}, \texttt{success}, \texttt{result} or \texttt{error} &
Associate feedback with the request. Errors contain \texttt{code}, \texttt{message},
and optional \texttt{details}. \\
\bottomrule
\end{tabular}
\end{table}

\paragraph{A complete invocation path.}
The following example illustrates a cleaning capability. It declares a string
argument \texttt{area}, an optional \texttt{mode} from
\texttt{auto}, \texttt{spot}, or \texttt{deep}, and an output containing
\texttt{accepted} and, when available, an \texttt{operation\_id}.
Its argument schema requires \texttt{area} and disallows undeclared
properties; the output schema requires the Boolean \texttt{accepted}.
The same operation is registered, invoked, and returned as follows:

\begin{verbatim}
# Register the declared schema and its device implementation.
registry.register(
    provider_id="vacuum_01",
    capability={
        "name": "cleaning.clean_area",
        "version": "1.0",
        "description": "Start cleaning a specified area.",
        "input_schema": CleanAreaInput,
        "output_schema": CleanAreaOutput
    },
    implementation=clean_area
)

# Invocation envelope produced for the selected operation.
{
  "schema_version": "0.1",
  "request_id": "req_1024",
  "provider_id": "vacuum_01",
  "capability": {"name": "cleaning.clean_area", "version": "1.0"},
  "arguments": {"area": "kitchen", "mode": "spot"}
}

# The registered function wraps the existing device controller.
def clean_area(area, mode="auto"):
    return vacuum_sdk.start_cleaning(
        target_area=area, cleaning_mode=mode
    )

# Return envelope: acceptance is distinct from task completion.
{
  "schema_version": "0.1",
  "request_id": "req_1024",
  "success": true,
  "result": {"accepted": true, "operation_id": "clean_381"}
}
\end{verbatim}

\paragraph{Validation, dispatch, and feedback.}
Before dispatch, the runtime checks provider availability, capability
membership, version compatibility, and argument-schema conformance.
The registered function executes through its underlying controller and
returns a schema-conforming result or a structured error.
For example, an unavailable provider returns \texttt{success: false}
with the code \texttt{PROVIDER\_UNAVAILABLE}; other errors distinguish
invalid arguments, denied permissions, unmet preconditions, execution
failure, and timeout. The Response Model receives this feedback in its
continuing context and can revise the response or select another
available operation.

A syntactically valid call does not itself grant permission or establish
physical completion. Execution limits belong to the provider and runtime;
task completion is established through feedback and subsequent observations.
Retries, replanning, and recovery are agent/runtime decisions, not implied
by the function signature. The callable boundary can use a local function,
an SDK wrapper, or a transport such as MCP, HTTP, ROS, or WebSocket without
changing the operation's declared input and output semantics.

\paragraph{From device operations to reusable skills.}
The same boundary supports a progression from device primitives to
bounded skills and compositions of skills. A specialized VLA could be
exposed as a callable skill: the high-level model selects a goal, while
the specialized policy maintains its own perception and low-level control.
Its capability description must specify the supported observations,
workspace, arguments, and termination conditions.

This separation also provides a route to harness-level skill evolution.
Execution feedback could motivate revised skill descriptions, parameters,
wrappers, or reusable compositions, with proposed updates tested before
deployment. These are extensions enabled by the architecture, rather than
additional capabilities established by the present experiments.
Section~\ref{app:engine-challenge-outcome-learning} develops the corresponding
learning and validation questions.

\section{Open Challenges and Research Directions}
\label{app:engine-open-challenges}

The modular Engine is one realization of ProPAI, not a restriction on
future architectures. The following questions describe a broader research
agenda connecting continuous world understanding, intervention learning,
and physical execution. Each box states an open problem; the accompanying
text outlines candidate approaches and the evidence needed to assess them.

\subsection{Persistent World Understanding and Memory}
\label{app:engine-challenge-memory}
\label{app:engine-challenge-world-models}

\begin{propaopenchallenge}[label={challenge:propa-memory}]
\label{challenge:propa-world-models}
How can a proactive system connect past commitments, present evidence,
and possible futures to recognize when intervention becomes useful?
\end{propaopenchallenge}

\paragraph{Remembering unresolved needs.}
Memory provides the starting point for persistent world understanding:
preserving task-relevant continuity without
confusing historical evidence with the present world. A task may remain
unresolved after its original visual evidence disappears, while a visible
hazard may already have been handled by someone else. A future memory
system could combine an event-and-entity representation with explicit task
commitments and intervention outcomes, retrieving information according
to its causal relevance rather than textual similarity alone.

Learning when to write, retrieve, consolidate, revise, and forget may be
as important as learning the response policy. User corrections should
update the relevant belief without silently changing unrelated memories;
uncertain identities and conflicting observations should remain explicit.
Evaluation could test resumption of interrupted tasks, redundant
interventions, errors caused by stale beliefs, and correction propagation
over extended sessions. Comparisons should control context and retrieval
budgets and include retention and deletion requirements, rather than
equating a larger memory with a better proactive system.

A concrete starting point is a layered memory: a short-lived perceptual
buffer for recent evidence, an episodic record of interventions and their
outcomes, and a persistent task state for commitments that outlive either
window. A task record could link its initiating evidence, responsible
person or device, completion condition, and next relevant observation.
For example, an illustrative cooking assistant should distinguish a pot
that still needs attention from one already removed by the user, even if
both situations follow the same earlier reminder. Retrieval would then
answer not only ``what happened?'' but ``what remains to be done?''

This organization suggests different memory access paths for the two
models. The Timing Model could consult a compact account of unresolved
conditions at each update, while the Response Model retrieves supporting
episodes only when planning assistance. The key question is which state
can be compressed without erasing the evidence needed to change a
decision. Paired histories with identical recent frames but different
prior commitments or completed actions would help isolate this ability
from short-window visual recognition.

\paragraph{Anticipating actionable changes.}
While memory tracks what remains unresolved, prediction connects subtle precursors to the consequences of waiting
or acting. For example, an illustrative system might estimate whether an
unstable object is likely to fall and whether a warning or available
device action could still help. The relevant prediction is not only what
the scene will look like, but how possible interventions change the
outcomes and the remaining time in which they are useful.

A future design could use a predictive model to propose response windows
or compare candidate interventions, while grounding the final decision in
observed evidence, available capabilities, and uncertainty. Forecasts
should remain distinguishable from observations in memory and planning.
Evaluation should separate predictive accuracy from intervention benefit:
an accurate forecast can still produce an unnecessary, infeasible, or late
response. Controlled scenarios and appropriately designed comparisons
would be needed to assess the benefit of anticipation rather than assume
it from visually plausible predictions.

Prediction could also be conditioned on the available executor. A warning
that a person can act on immediately and a robot action that requires
navigation have different useful starting times. Rather than optimizing
forecasting in isolation, a future system could estimate the remaining
opportunity for each feasible response and update that estimate as new
evidence arrives. Comparisons between observation-only and predictive
policies should therefore examine both interventions made possible by
earlier preparation and false interventions induced by uncertain forecasts.

\subsection{Native Proactive Models and Real-Time Computation}
\label{app:engine-challenge-native}
\label{app:engine-challenge-efficiency}

\begin{propaopenchallenge}[label={challenge:propa-native}]
\label{challenge:propa-efficiency}
How can a model jointly learn continuous observation, intervention timing,
and response generation while allocating computation to meet intervention deadlines?
\end{propaopenchallenge}

\paragraph{Joint modeling of timing and response.}
A native model could internalize some of the coordination currently
provided by the Timing Model, Response Model, and Harness. Candidate
architectures include a shared streaming backbone with specialized
decision and response pathways, or a persistent recurrent state that
supports both intervention decisions and language or capability outputs.
The scientific question is whether shared representations improve the
coupling between timing and response without allowing expensive generation
to delay observation updates or degrade restraint.

Such a model would need supervision for silence, response initiation,
continuation, and closure alongside response content and execution
feedback. Silence is not simply the absence of an answer token: it is a
decision made while observations continue to arrive. A fair comparison
with the modular system should match data, modalities, context, and compute,
and assess both response quality and decision latency. Unification would
not remove the need for external device interfaces, permission enforcement,
or independently observable execution outcomes.

One architectural route is to maintain a shared causal state with a
lightweight decision pathway and a selectively activated response pathway.
Observation updates would continue during response generation, allowing
new evidence to revise an unfinished plan before it is dispatched.
This raises a concrete state-consistency problem: the system must know
which observations support a draft response and when later evidence
invalidates it. Another route is to distill coordinated behavior from
the modular system into a native model, then extend training to continuous
interaction sequences rather than isolated question--answer examples.
The modular design provides a useful experimental reference because
timing, generation, and execution feedback can be varied separately.

\paragraph{Adaptive computation.}
Incremental Caching, Dynamic Token Dropping, and Prefix-constrained Decoding address local
sources of overhead. A broader question is how much computation each
moment deserves. Slowly changing scenes may permit inexpensive monitoring,
whereas an ambiguous change near a response boundary may justify denser
observations, retained visual detail, or escalation to a larger model.
Candidate policies could adapt observation cadence, token retention,
retrieval depth, and response preparation to uncertainty and estimated
urgency. They must also preserve coverage of unexpected events that a
cheap monitor might otherwise fail to notice.

This creates a joint timing and resource-allocation problem. Useful
comparisons would report the trade-off among intervention quality, tail
latency, energy, and memory under the same event stream and external
deadlines. Stress tests could include simultaneous tasks, network delays,
long histories, and bursts of sensor input. A configuration should not
be called more efficient merely because it silently skips difficult
observations or reduces the opportunities at which a response is evaluated.

In particular, efficient inference should be studied as a scheduling
problem as well as a model-compression problem. Observation encoding,
memory retrieval, speculative response preparation, and output decoding
compete for a finite compute budget. A scheduler could prioritize a new
high-urgency observation over an obsolete draft, or reuse prepared context
when the triggering evidence remains valid. Experiments should charge
discarded speculative work to the compute budget and count missed
opportunities during busy periods. This would distinguish genuine gains
from shifting latency into an unreported queue or precomputation phase.

\subsection{From Execution Feedback to Capability Growth}
\label{app:engine-challenge-outcome-learning}
\label{app:engine-challenge-vla}
\label{app:engine-challenge-self-evolution}

\begin{propaopenchallenge}[label={challenge:propa-outcome-learning}]
\label{challenge:propa-vla}
\label{challenge:propa-self-evolution}
How can experience improve intervention decisions, expand the set of
executable skills, and refine their composition into reliable assistance?
\end{propaopenchallenge}

\paragraph{Learning from preferences and outcomes.}
Complete events suggest useful comparisons: a timely intervention may be
preferable to a late one, restraint to an unnecessary interruption, and a
feasible response to an unavailable action. These comparisons are conditional
on the task and environment rather than universal reward rules. Future
work could investigate temporal preference learning or outcome-based
reinforcement learning, using verified execution traces and human
corrections to complement broad video supervision.

Outcome attribution is a central difficulty. An observed recovery may be
caused by another person, by the environment itself, or by the system's
action; a failure may arise from the timing decision, the plan, or the
executor. Moreover, logs produced by one policy do not reveal what would
have happened under every alternative. Progress requires preserving
provenance and distinguishing these sources of uncertainty. Evaluation
should use held-out tasks and independent outcome checks, and retain
unnecessary interventions and failures rather than training only from
successful trajectories. 

A useful progression would begin with matched timing preferences under
the same context, then introduce closed-loop episodes in which an
intervention changes later observations. Timing and response could first
be optimized separately with the other component fixed, before exploring
joint learning. This staged comparison would help determine whether a
gain comes from recognizing the need earlier, selecting a better action,
or executing it more reliably. Feedback should retain these distinctions:
a fluent response, an accepted device request, and a completed physical
task are different learning signals. Where counterfactual outcomes are
unavailable, preference judgments and observed outcomes should remain
separate rather than being treated as interchangeable rewards.

\paragraph{Specialized skills and cross-embodiment composition.}
Exposing specialized VLAs as callable functions would let a high-level model reason
about goals while each executor handles its local perception and control.
The difficult part is establishing a useful capability contract: supported
objects, workspace, required observations, expected failure modes, and
conditions under which the policy should decline or stop. Capability
selection could depend on such information rather than treating every
executor as an interchangeable implementation of an abstract command.

An expanding skill library also changes which interventions are feasible
and how early they must begin. A slower but more reliable executor may
require a different response window; losing a manipulation skill may make
a human warning preferable to direct action. A meaningful transfer study
would freeze the proactive models, vary the declared skill inventory or
embodiment, and measure capability selection, task completion, timing,
and failures under the changed conditions. Zero-shot substitution should
be distinguished from recalibration, wrapper engineering, or policy
retraining. A callable-function boundary makes this interface study possible.

For example, a manipulation policy could be exposed as a bounded
\texttt{pick\_and\_place} skill whose inputs specify a supported object
and destination. The VLA would consume its own visual and proprioceptive
observations and generate motor commands; the proactive model would
decide whether moving the object is useful, choose an eligible executor,
and interpret progress or failure. The wrapper would translate task-level
arguments into the policy's native interface and return explicit status,
including whether execution can be interrupted. This division connects
high-level initiative to specialized control without requiring the
proactive model to generate motor actions itself.

Composition adds another question: when does one skill's result satisfy
the next skill's preconditions? A navigation--inspection--manipulation
sequence may require fresh observations between stages rather than an
unconditional list of calls. Skill interfaces could expose these
dependencies, resource requirements, and completion evidence. Testing
the same high-level task with different skill inventories would reveal
whether the model adapts its plan, seeks human assistance, or appropriately
refrains when the required capability is absent.

\paragraph{Harness-level skill evolution.}
Once skills are exposed through a common interface, their selection and
composition become another locus of learning. Self-evolution need not
begin with changing the foundation model's weights.
A Harness could propose improved capability descriptions, retrieval rules,
tool-selection prompts, wrappers, or reusable compositions of existing
skills. Repeated failures might expose a missing precondition; repeated
successful action sequences might motivate a candidate composite skill.
The crucial distinction is between proposing a change and deploying it.
Generating a new function name does not create a new physical ability,
and a successful trace does not establish general correctness.

A candidate development loop would retain the triggering evidence,
generate a versioned proposal, test it on recorded cases and independent
held-out scenarios, and evaluate it in a sandbox or suitable simulator
before any authorized physical trial. Promotion to the live registry would
require explicit acceptance criteria, compatibility checks, and a rollback
path. Changes that affect physical behavior or permissions would require
the corresponding authorization. The evolving component should not be
allowed to redefine the evaluation criteria, grant itself new permissions,
or disable the safeguards used to assess it.

The unit of improvement can be made explicit. Repeated selection errors
may motivate a clearer capability description; a recurring argument
failure may motivate a wrapper check; a successful multi-step procedure
may motivate a reusable composite skill. Each proposal should identify
the failure pattern it addresses and preserve cases where the previous
version already worked. An illustrative inspection routine, for example,
could evolve to request a second view when the first is occluded, rather
than treating every uncertain observation as a negative finding.
Such an update changes the evidence-gathering procedure, not necessarily
the foundation model or physical controller.

Evaluation should separate gains from additional memory, better routing,
new compositions, and genuinely improved executors. A frozen base model
with fixed evaluation tasks would provide one way to isolate improvements
in the Harness itself. Relevant outcomes include transfer beyond the
triggering failure cases, regression frequency, test and deployment cost,
and recovery after an accepted change later proves unsuitable. 

\subsection{Long-Term Human and Multi-Actor Collaboration}
\label{app:engine-challenge-agency}

\begin{propaopenchallenge}[label={challenge:propa-agency}]

How should an always-on system coordinate evolving needs, shared devices,
and human preferences while preserving people's control over intervention?
\end{propaopenchallenge}

Proactivity creates obligations beyond executing a requested command.
An intervention may interrupt one person to help another, occupy a shared
device, or reveal information that should remain private. A future
coordination layer could represent task urgency, resource commitments,
explicit user preferences, and authorization requirements without treating
inferred personal attributes as permission. Uncertain preferences may
call for clarification, while withdrawal of consent should change the
allowed behavior even if a task remains technically feasible.

Coordination also requires an account of task ownership. A person may
already be addressing a problem that a robot has just detected, or two
devices may offer overlapping capabilities. A future system could track
who has acknowledged a task, which resources are committed, and what
evidence would justify a handoff or renewed reminder. Communication is
then part of coordination rather than merely the final output: a brief
clarification can resolve ambiguous responsibility before an unnecessary
action is taken. Personalization could adjust reminder frequency or
preferred assistance channels while keeping explicit instructions distinct
from preferences inferred from sparse behavior.

Evaluation should include overlapping needs, conflicting requests,
unavailable devices, changing permissions, human takeover, and situations
where no intervention is appropriate. Task completion alone cannot capture
unnecessary interruption, privacy violations, or failure to relinquish
control. Longer-term studies could examine whether assistance remains
useful after personalization and repeated exposure, while keeping data
retention and correction under user control. These questions extend embodied evaluation beyond a fixed scenario inventory
to sustained collaboration in changing environments.

Longitudinal evaluation would be particularly informative here. The same
participants could encounter recurring and novel situations across
sessions, with changing responsibilities and deliberate task handoffs.
Besides completion and response timing, useful measurements include
duplicate interventions, successful handoffs, corrections needed from
users, and the persistence of outdated preferences. Comparing a fixed
policy with memory-only and adaptive variants would help separate the
benefits of remembering past interactions from those of changing the
intervention policy itself.

\label{app:engine-agenda-synthesis}
These directions interact rather than forming independent additions.
Memory changes which events appear new or unresolved; better prediction
changes when evidence becomes actionable; new skills change what can be
done; and self-evolution changes the procedures used to select and compose
those skills. A native model may internalize some of these relationships,
while an explicit Harness keeps their effects inspectable. The unifying
research question is how to improve the complete path from an emerging
need to a timely, appropriate, and verifiable response without losing
causal grounding or human control.

\endgroup

\end{document}